\documentclass{article} 
\usepackage{iclr2024_conference,times}

\usepackage{amsmath,amsfonts,bm}

\def\eqref#1{equation~\ref{#1}}

\def\1{\bm{1}}

\DeclareMathAlphabet{\mathsfit}{\encodingdefault}{\sfdefault}{m}{sl}
\SetMathAlphabet{\mathsfit}{bold}{\encodingdefault}{\sfdefault}{bx}{n}

\usepackage{hyperref}
\usepackage{url}
\usepackage{multicol}
\usepackage{multirow}
\usepackage{graphicx}
\usepackage[section]{placeins}
\usepackage{multirow}
\usepackage{graphicx} 
\usepackage{amsmath} 
\usepackage{siunitx} 
\usepackage{colortbl}
\usepackage{xcolor}
\definecolor{lightgray}{gray}{0.94}
\definecolor{mypurple}{RGB}{112, 48, 160}
\definecolor{myyellow}{RGB}{234, 234, 0}
\definecolor{darkpink}{rgb}{0.91, 0.33, 0.5}
\usepackage{booktabs}
\usepackage{makecell}
\usepackage{authblk}
\title{PEAR: Phrase-Based Hand-Object Interaction Anticipation}

\author[1]{Zichen Zhang}
\author[2]{Honhchen Luo}
\author[1]{Wei Zhai}
\author[1,3]{Yang Cao}
\author[1,3]{Yu Kang}
\affil[1]{University of Science and Technology of China, Hefei, China}
\affil[2]{Northeastern University, Shenyang, China}
\affil[3]{Institute of Artificial Intelligence, Hefei Comprehensive National Science Center, Hefei, China}
\iclrfinalcopy 
\begin{document}

\maketitle

\begin{abstract}
First-person hand-object interaction anticipation aims to predict the interaction process over a forthcoming period based on current scenes and prompts. 
This capability is crucial for embodied intelligence and human-robot collaboration. 
The complete interaction process involves both pre-contact interaction intention (\textit{i.e.}, hand motion trends and interaction hotspots) and post-contact interaction manipulation (\textit{i.e.}, manipulation trajectories and hand poses with contact). 
Existing research typically anticipates only interaction intention while neglecting manipulation, resulting in incomplete predictions and an increased likelihood of intention errors due to the lack of manipulation constraints.
To address this, we propose a novel model, \textbf{PEAR} (Phrase-Based Hand-Object Interaction Anticipation), which jointly anticipates interaction intention and manipulation. 
To handle uncertainties in the interaction process, we employ a twofold approach.
Firstly, we perform cross-alignment of verbs, nouns, and images to reduce the diversity of hand movement patterns and object functional attributes, thereby mitigating intention uncertainty. 
Secondly, we establish bidirectional constraints between intention and manipulation using dynamic integration and residual connections, ensuring consistency among elements and thus overcoming manipulation uncertainty.
To rigorously evaluate the performance of the proposed model, we collect a new task-relevant dataset, \textbf{EGO-HOIP}, with comprehensive annotations. 
Extensive experimental results demonstrate the superiority of our method.
\end{abstract}

\section{Introduction}
\begin{figure}[h]
\centering
\includegraphics[width=0.88\linewidth]{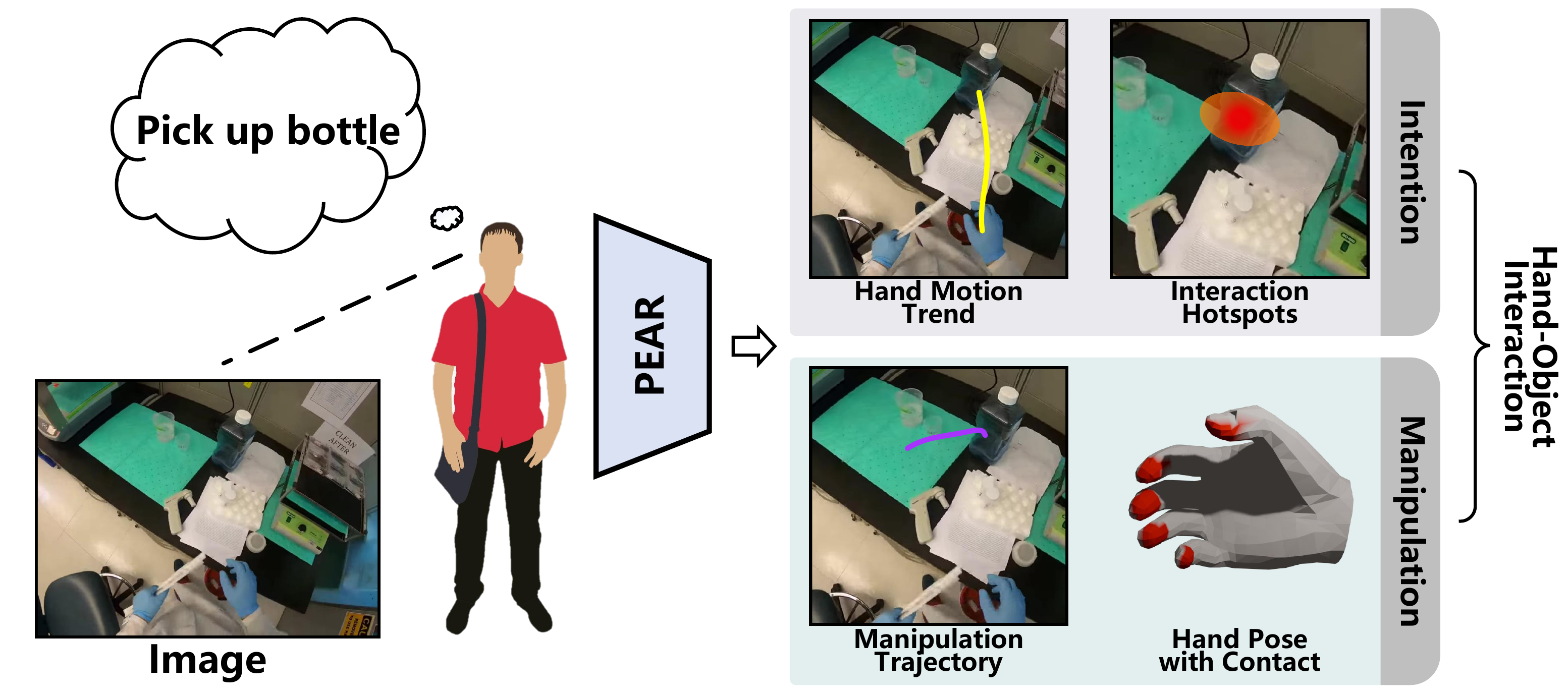}
\caption{Given an image of the pre-interaction scenario and a phrase, \textbf{PEAR} anticipates the hand-object interaction process over a period of time, including interaction intention (\textit{i.e.}, hand motion trends and interaction hotspots) and interaction manipulation (\textit{i.e.}, manipulation trajectories and hand poses with contact).}
\label{fig:intro_simple1}
\end{figure}

Hands serve as the primary medium through which humans interact with the physical world.
Understanding and anticipating hand-object interactions can enable intelligent agents to learn and emulate human skills~\cite{skill1, skill2, skill3}.
The development of various head-mounted devices has increasingly facilitated the capture of first-person natural human interaction behaviors, providing robust data support for hand-object interaction research~\cite{egtea, epic, epic2, ego4d, ego-exo}.
First-person hand-object interaction anticipation aims to accurately and meticulously analyze future interaction processes based on given interaction scenarios and prompts. 
This includes hand movement planning before contacting objects, reflecting interaction intention, and the manner of operation after contact, reflecting interaction manipulation. 
This research has potential applications in fields such as AR/VR~\cite{AR1, AR2, AR3, VR1, vr2}, imitation learning~\cite{imi1, imi2, imi3}, human-robot collaboration~\cite{co1, co2} and embodied intelligence~\cite{emai1, emai2, emai3, emai4}.

\textit{How can machines anticipate the hand-object interaction?} 
One potential solution involves predicting specific elements that are capable of revealing the interaction. 
Existing research primarily focuses on predicting one or more intention elements to analyze hand-object interaction at a future moment or over a period of time.
To guide hand movement trends, some studies~\cite{aff1, aff2} forecast contact areas on objects based on videos, images, or text, thereby providing intelligent agents with knowledge of \textit{``where to touch''}. 
However, interaction hotspots alone are insufficient for detailed hand movement planning. 
To overcome this limitation, several studies~\cite{3dtraj, both1} forecast hand trajectories over time, offering more detailed motion planning for intelligent agents.
Building on these, other studies~\cite{both2, zzc} further leverage the constraints between object functionalities and hand movement patterns to jointly predict interaction hotspots and hand motion trends before hand-object contact, thereby improving the accuracy and coherence of interaction intention predictions.
Nevertheless, these studies do not consider manipulation elements, which reduces the completeness of the predictions and fails to provide guidance on \textit{``how to manipulate''} for intelligent agents.
Additionally, since intention and manipulation together constitute a comprehensive interaction process, the absence of manipulation elements increases the likelihood of errors in anticipating interaction intention.

\begin{figure}[t]
\centering
\includegraphics[width=1\linewidth]{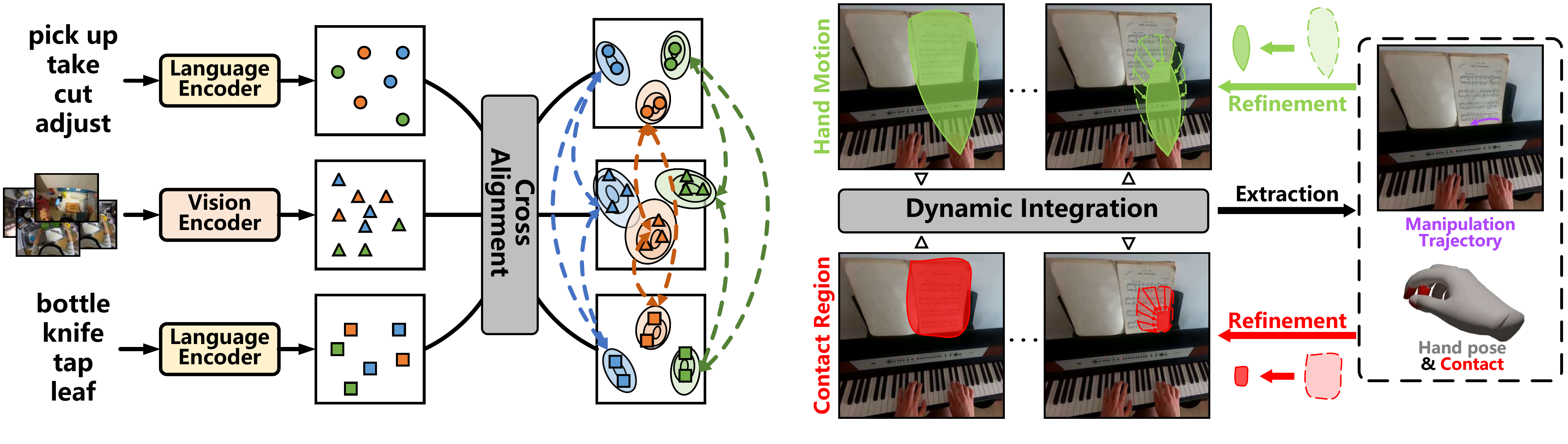}
    \put(-298,-4){\makebox(0,0){\textbf{(a)}}}
    \put(-91,-4){\makebox(0,0){\textbf{(b)}}}
\caption{\textbf{Motivation.} We address both intention uncertainty and manipulation uncertainty with specific solutions. \textbf{(a)} We reduce the scope of intention elements by cross-aligning nouns, verbs, and images, thereby overcoming intention uncertainty. \textbf{(b)} We derive manipulation elements through the dynamic integration of intention elements, simultaneously refining the initial intention via manipulation, thereby mitigating manipulation uncertainty.}
\label{motivation}
\end{figure}

To address these challenges, we propose to leverage phrases and images as inputs to jointly predict intention elements prior to hand-object contact (including hand motion trends and interaction hotspots) and post-contact manipulation elements (including manipulation trajectories and hand poses with contact), as demonstrated in Fig.~\ref{fig:intro_simple1}.
However, anticipating complete and accurate hand-object interaction is challenging due to the inherent uncertainty in the interaction process.
Firstly, hands exhibit high flexibility and low spatial constraints, leading to various movement patterns before touching objects.
Meanwhile, objects possess multiple functional attributes, resulting in diverse contact areas.
The combined diversity of hand movements and object functionalities leads to \textbf{intention uncertainty}. 
To mitigate this, we leverage the relationships between nouns and object functionalities, verbs and movement patterns, and the pairing of nouns and verbs to cross-align nouns, verbs, and images. This approach reduces the intention uncertainty, as illustrated in Fig.~\ref{motivation}\textbf{(a)}.
Secondly, interaction manipulation occurs in the latter part of the interaction process and is highly sensitive to variations in intention. 
Both hand motion trends and interaction hotspots span a certain range and exhibit a large gap at the feature level.
Consequently, establishing a stable and accurate matching relationship between them is difficult, leading to imprecise extraction of post-contact elements and resulting in \textbf{manipulation uncertainty}.
To address this, we introduce a dynamic bidirectional constraint relationship between intention and manipulation based on their shared interaction logic.
This approach narrows the range of intention elements while extracting manipulation elements, thereby ensuring the continuity and consistency of manipulation with respect to intention and mitigating manipulation uncertainty, as shown in Fig.~\ref{motivation}\textbf{(b)}.

To achieve this, we propose \textbf{PEAR} (\textbf{P}hrase-bas\textbf{E}d H\textbf{A}nd-Object Inte\textbf{R}action Anticipation), a novel phrase-based model that jointly anticipates both pre-contact and post-contact interaction elements. To overcome intention uncertainty, we employ image-text matching models as backbones to extract hand movement patterns and object functional attributes from interaction prompts and scenarios.
Parallel cross-attention modules are introduced to establish constraint relationships within intention elements, narrowing their respective ranges.
To tackle manipulation uncertainty, we propose a joint prediction framework based on a Deep Equilibrium extraction model, which dynamically resolves the fixed point between two intention elements to extract accurate manipulation elements. 
Residual connection blocks are further established from manipulation to intention, thereby enhancing the overall coherence and consistency of the predictions.
Additionally, to capture the natural randomness of human behaviors, we utilize the C-VAE as the decoder for each interaction element, introducing small, reasonable variations to the anticipated results.

Due to the lack of task-relevant datasets, we collect a new \textbf{EGO-HOIP} dataset to better validate PEAR's effectiveness for the proposed task.
This dataset comprises 5k pairs of interaction scenarios and prompts, featuring 90 verb labels and 600 noun categories, and we generate over 15k high-quality annotations for all interaction elements through a combination of automatic and manual methods, offering strong support to thoroughly assess the network's generalization and robustness. 
PEAR achieves state-of-the-art results on EGO-HOIP, demonstrating its superiority in hand-object interaction anticipation. 

The main contributions of this paper are summarized as follows:

\textbf{1)} This paper thoroughly exploits the correlation among interaction counterparts to jointly predict interaction intention and manipulation, providing comprehensive interaction elements to anticipate hand-object interaction.

\textbf{2)} We present a novel model, PEAR, to anticipate the entire hand-object interaction process. By establishing dynamic, bidirectional constraint relationships among various interaction elements, PEAR effectively overcomes uncertainties in both intention and manipulation, thereby improving the comprehensiveness and precision of predictions.

\textbf{3)} A new task-relevant EGO-HOIP dataset is introduced, encompassing extensive interaction data and annotations, offering valuable support for hand-object interaction anticipation. Extensive experiments demonstrate the superiority of the proposed network.

\section{Related Work}
\subsection{Hand-Object Interaction Understanding}
Hand-object interaction understanding aims to comprehend and perceive the current hand-object interaction content. 
Various methods, such as 3D-CNNs~\cite{3DCNN1,3DCNN2,3DCNN3}, and Transformers~\cite{trans1,trans2,trans3}, are employed to semantically summarize the interaction from video sequences, thereby forming the foundation of this task. 
However, while these methods aid agents in grasping the logic of human behaviors, they encounter difficulties in analyzing detailed interaction elements. 
To address this limitation, some research shifted focus toward 3D hand pose estimation during hand-object interaction.

Early studies concentrated on single-hand pose estimation from RGB images, employing either parametric~\cite{npa1,npa2} or non-parametric~\cite{pa1,pa2} methods to regress the vertices of the 3D hand mesh. 
The parametric approach, primarily based on the MANO model~\cite{mano}, indirectly estimates 3D hand poses by regressing hand pose and shape parameters, offering robustness in scenarios involving occlusion and blur. 
The non-parametric approach directly regresses all hand mesh vertices for better object alignment, though it may distort hand poses under occlusion.
Recent research explored more complex scenarios, such as reconstructing interactions between two hands~\cite{2hand1,2hand2,2hand3} and investigating the simultaneous reconstruction of hand-held objects and hand poses based on known object models~\cite{6d1,6d2,6d3}.
Supported by diffusion models~\cite{diffusion}, several studies~\cite{dipo1,dipo2,dipo3} relaxed input conditions by extracting feature representations of hands and objects in latent space from RGB images and aligning them for direct reconstruction of hand as well as object poses.

While these methods advance agents' understanding from high-dimensional interaction semantics to detailed interaction elements, they are inadequate for assisting agents in planning future actions or predicting human intention, thus limiting effective human-robot coordination. This paper introduces a method that extends hand-object interaction understanding to anticipate future interaction details, thereby overcoming these issues.

\subsection{Hand-Object Interaction Prediction}
Hand-object interaction prediction aims to forecast future interactions based on current scenes and prompts. 
Early research primarily focused on action anticipation, predicting the semantic label of the next action~\cite{sta1,sta2,sta3,sta4} or a sequence of semantic labels~\cite{lta1,lta2,lta3} by analyzing previous actions.
However, their predictions remain at a semantic level, lacking detailed interaction specifics, thus failing to provide effective low-dimensional action planning guidance.

To address this, several studies predicted specific interaction intention elements, such as hand motion trends~\cite{3dtraj} and interaction hotspots~\cite{aff1, aff2, aff4, aff5, aff6}.
Based on this, some research leveraged the constraint relationships between these two elements to jointly anticipate interaction intention.
Liu et al.~\cite{both1} were the first to jointly predict hand motion trends and interaction hotspots.
Liu et al.~\cite{both2} improved the accuracy of joint predictions by incorporating the guiding influence of objects on hand movements.
Zhang et al.~\cite{zzc} further enhanced this by introducing a bidirectional progressive mechanism to exploit the mutual constraints between hand motion and object affordance.

However, these studies primarily focus on predicting interaction up to hand-object contact, neglecting the detailed analysis of post-contact manipulation.
Consequently, they fall short of providing a comprehensive prediction of hand-object interaction.
To overcome this, this paper introduced a novel model to anticipate both interaction intention and manipulation, by leveraging the shared interaction logic between intention and manipulation, the proposed approach mitigates inherent uncertainty in the interaction process by establishing bidirectional constraints among different interaction elements, thereby enhancing the accuracy of predictions.

\subsection{Hand-Object Interaction Generation}
Hand-object interaction generation aims to generate appropriate hand-object interaction content based on specific prompts. 
Initial research~\cite{mogen1,mogen2,mogen3,mogen4} focused on generating hand grasps, which involves creating a 3D hand pose interacting with a given object model. 
However, these methods were limited to static grasp poses and did not support the generation of dynamic motion sequences.
To address these limitations, Cha et al.~\cite{seqgen2} utilized prompts from textual descriptions to generate hand-object interaction motion sequences based on a provided 3D object model. 
Despite advancements, these methods were implemented in simulated environments, lacking real-world challenges like occlusion and blur, resulting in relatively simplistic action sequences that are difficult to apply directly to real-world scenarios.

To address these limitations, several studies progressed toward generating 3D hand poses based on real interaction scenarios. 
Corona et al.~\cite{rgbgen1} developed a method to predict how a human would naturally grasp one or several objects given a single RGB image of these objects. 
Narasimhaswamy et al.~\cite{handiffuser} explored generating hand-object interaction images in real-world contexts using only textual descriptions.
Compared to general generation methods, these approaches ensured that the generated hands and objects adhered to natural logic and maintained a high degree of congruence. 
Nevertheless, the generated interaction content remained confined to static scenes, without incorporating dynamic processes.

Indeed, generating complete hand-object interaction motion sequences in real-world scenarios presents significant challenges, particularly in ensuring the plausibility and detailed accuracy of the interaction.
Consequently, the method proposed in this paper does not attempt to generate full 3D interaction sequences. 
Instead, it capitalizes on the characteristic of less pose variation during object manipulation. By focusing on predicting the hand pose at the moment of contact and the subsequent manipulation trajectory, this approach offers practical and accurate guidance for manipulation. 

\section{Method}
\begin{figure}
    \centering
    \includegraphics[width=1\linewidth]{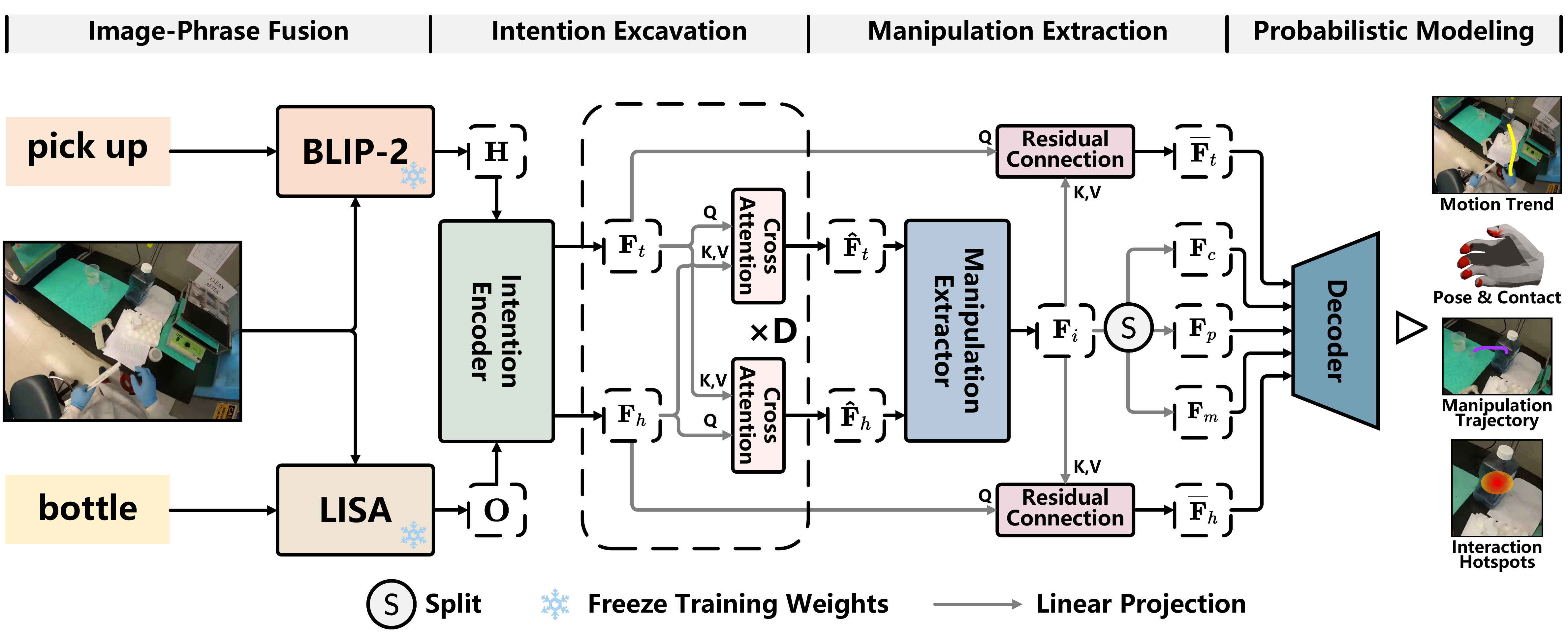}
    \caption{\textbf{PEAR pipeline}. The proposed model takes an image and a phrase as inputs to anticipate future interaction elements. It consists of four components: an image-phrase fusion module, an interaction intention excavation module, an interaction manipulation extraction module, and a probabilistic modeling prediction module.}
    \label{method}
\end{figure}

Given a pre-interaction image $\mathcal{I}$ as the interaction scenario and a phrase $\mathcal{E}$ as the interaction prompt, the proposed method anticipates the complete hand-object interaction process, which includes interaction intention (hand motion trends $\mathcal{T}$ and interaction hotspots $\mathcal{O}$) and interaction manipulation (manipulation trajectories $\mathcal{M}$ and hand poses $\mathcal{P}$ with contact $\mathcal{C}$). As shown in Fig.~\ref{method},
initially, in the \textbf{Image-Phrase Fusion Module} (Sec.~\ref{sec3.1}), the phrase is divided into a verb $\mathcal{E_V} $ and a noun $\mathcal{E_N} $, which are then fused with the original image $\mathcal{I}$ via dual branches, producing the hand motion feature $\mathbf{H}$ and the contact area feature $\mathbf{O}$. However, $\mathbf{H}$ and $\mathbf{O}$ remain ambiguous and lack inter-element constraints. 
To overcome intention uncertainty, we introduce the \textbf{Interaction Intention Excavation Module} (Sec.~\ref{sec3.2}). In this module, $\mathbf{H}$ and $\mathbf{O}$ are refined through an intention extraction encoder, converting the intention elements into the more structured hand motion feature $\mathbf{F}_t$ and interaction hotspots feature $\mathbf{F}_h$. On this basis, parallel cross-attention blocks are established between $\mathbf{F}_t$ and $\mathbf{F}_h$, creating a bidirectional constraint relationship that further specifies the interaction intention, yielding $\mathbf{\hat F}_{t}$ and $\mathbf{\hat F}_{h}$.
Subsequently, to mitigate manipulation uncertainty, an \textbf{Interaction Manipulation Extraction Module} (Sec.~\ref{sec3.3}) is employed,  using an ``infinite-depth'' DEQ extraction model to derive the manipulation feature $\mathbf{F}_i$ from $\mathbf{\hat F}_{t}$ and $\mathbf{\hat F}_{h}$. Two residual connection modules are established between $\mathbf{F}_i$ and $\mathbf{F}_t$, as well as $\mathbf{F}_i$ and $\mathbf{F}_h$, refining the initial interaction intention. $\mathbf{F}_i$ is further decomposed into the manipulation trajectory feature $\mathbf{F}_m$, hand pose feature $\mathbf{F}_p$, and hand contact feature $\mathbf{F}_c$, thus completing the extraction of all interaction element features.
Eventually, C-VAEs are utilized as the backbone to design the \textbf{Probabilistic Modeling Decoder} (Sec.~\ref{sec3.4}), anticipating each interaction element while replicating the natural and reasonable randomness of human behaviours. The whole process is optimized by a combined loss (Sec.~\ref{sec3.5}).

\subsection{Image-Phrase Fusion Module}
\label{sec3.1}
While methods such as CLIP~\cite{clip} and BLIP~\cite{blip} are proficient in tasks like image-text matching, their capabilities are primarily limited to comprehending the current scene. 
These approaches encounter difficulties in directly translating text-implied actions into specific visual representations, such as hand movement trajectories and interaction hotspots.
Typically, in textual descriptions, nouns denote attributes such as object locations and shapes, whereas verbs encapsulate patterns of hand motion.
To advance beyond merely understanding the image-text scene toward action-scenario matching, our approach separates nouns and verbs within the Image-Phrase Fusion Module, aligning and integrating them with the original image independently.
In the first branch, the verb $\mathcal{E_V}$ and the image $\mathcal{I}$ are processed through a pre-trained BLIP-2 model~\cite{blip-2} to extract the hand motion feature $\mathbf{H}$. 
Concurrently, in the second branch, the noun $\mathcal{E_N}$ and the image $\mathcal{I}$ are jointly processed by the pre-trained image-text segmentation model, $\mathrm{LISA}$~\cite{lisa}, yielding an object-level contact area feature $\mathbf{O}$.
By differentiating the processing of the verb $\mathcal{E_V}$ and the noun $\mathcal{E_N}$, the proposed approach facilitates more precise alignment and integration of visual and textual inputs, which is crucial for advancing from simple scene comprehension to action planning and interaction scenario matching.

\subsection{Interaction Intention Excavation Module}
\label{sec3.2}
Although $\mathbf{H}$ and $\mathbf{O}$ exhibit regularities pertinent to interaction scenarios and prompts, the hand motion feature $\mathbf{H}$ remains imprecise when projected onto specific scenarios, and the contact area feature $\mathbf{O}$ only provides a general object location, lacking guidance for specific contact regions. To overcome this intention uncertainty, we first employ an intention encoder to refine $\mathbf{H}$ and $\mathbf{O}$.
Specifically, self-attention blocks~\cite{attention} are employed to extract a more specific motion feature $\mathbf{F}_t$, thereby narrowing the range of hand motion trends. 
The i-th block can be expressed as:
\begin{equation}
    \mathbf{y}_i = \mathbf{x}_{i-1} + \mathrm{MSA}(\mathrm{LN}(\mathbf{x}_{i-1})), \quad i \in [0, n],
\end{equation}
\begin{equation}
    \mathbf{x}_i = \mathbf{y}_i + \mathrm{FFN}(\mathrm{LN}(\mathbf{y}_i)),
\end{equation}
where $\mathbf{x} _{0}$ represents the original hand motion pattern feature $\mathbf{H}$, $\mathbf{x} _{n}$ indicates the corrected feature $\overline{\mathbf{H}}$,
$\mathrm{MSA}$ stands for Multi-head Self-Attention, query, key, and value are processed through distinct linear layers: $\mathrm{MSA}(\mathrm{X}) = \mathrm{Attention}(\mathbf{W}^Q \mathrm{X}, \mathbf{W}^K \mathrm{X}, \mathbf{W}^V \mathrm{X}).$
$\mathrm{FFN}$ refers to the Feed Forward Layer, comprising a two-layer $\mathrm{MLP}$, and Layer normalization ($\mathrm{LN}$) is applied before each block.
Meanwhile, an image encoder (Segformer-b2~\cite{segformer}) is introduced to extract a more detailed contact area feature $\mathbf{F}_h$.
Additionally, to establish preliminary bidirectional constraints between $\mathbf{F}_t$ and $\mathbf{F}_h$, parallel cross-attention blocks with a depth $\mathbf{D}$ are introduced between these two components, resulting in updated features $\mathbf{\hat F}_{t}$ and $\mathbf{\hat F}_{h}$. This process completes the alignment and fusion of intention elements, which can be formulated as:
\begin{equation}
    \mathbf{\hat F}_{t} = \mathrm{MCA}(\mathbf{F}_t, \mathbf{F}_h), \quad 
    \mathbf{\hat F}_{h} = \mathrm{MCA}(\mathbf{F}_h, \mathbf{F}_t),
\end{equation}
where $\mathrm{MCA}$ stands for Multi-head Cross-Attention, query, key, and value are processed through linear layers: ${\mathrm{MCA} (\mathrm{X,Y} ) = \mathrm{Attention}(\mathbf{W} ^{Q} \mathrm{X} ,\mathbf{W} ^{{K} } \mathrm{Y},\mathbf{W} ^{V} \mathrm{Y} )}$.
In this module, the mutual guidance relationship between intention elements helps mitigate the intention uncertainty, providing more accurate initial conditions for the extraction of subsequent manipulation elements.

\subsection{Interaction Manipulation Extraction Module}
\label{sec3.3}
After obtaining the intention features, to address manipulation uncertainty, it is imperative to perform dynamic deep alignment and fusion of $\mathbf{\hat F}_{t}$ and $\mathbf{\hat F}_{h}$ to ensure that the extracted manipulation feature is both stable and accurate.
Traditional methods, mainly based on CNNs ~\cite{resnet,scis_cnn} and Transformers~\cite{attention, scis_trans}, operate with limited and fixed network depths. When dealing with dynamic input features (\textit{e.g.}, hand motion feature), these networks are unable to adjust their depth to achieve a fixed point, making it challenging to attain a stable equilibrium state between different features. To address this limitation, we introduce a deep equilibrium-based feature extraction framework. By simulating ``infinite'' depth, this framework dynamically solves for unified features between $\mathbf{\hat F}_t$ and $\mathbf{\hat F}_h$, thereby extracting the precise manipulation feature $\mathbf{F}_i$.
A standard Deep Equilibrium model~\cite{deq} follows the formula given below:
\begin{equation}
    \mathbf{z}^0 = 0, \quad \mathbf{z}^{k+1} = \mathit{f}_\theta (\mathbf{z}^k, \mathbf{x}),
\end{equation}
where the input $\mathbf{x}$ is injected at each layer by the residual block $\mathit{f} _{\theta }$, and $\mathbf{z} ^{k}$ is the hidden feature at layer $k$.
As the depth approaches infinity, the state $\mathbf{z}$ reaches equilibrium $\mathbf{z} ^* = \mathit{f} _\theta (\mathbf{z} ^*,\mathbf{x} )$.
To accelerate the convergence process, the DEQ model typically reformulates the equilibrium finding as a root-finding problem~\cite{deq}:
\begin{equation}
    \mathit{g}_\theta(\mathbf{z}, \mathbf{x}) = \mathit{f}_\theta (\mathbf{z}, \mathbf{x}) - \mathbf{z}, \quad \mathbf{z}^* = \mathrm{RootSolver}(\mathit{g}_\theta, \mathbf{x}),
    \label{eq11}
\end{equation}
where $\mathit{g} _\theta$ is the function representing the residual of the equilibrium equation.
In the proposed module, we employ the DEQ extraction model~\cite{deqfusion} to achieve deep alignment and fusion of $\mathbf{\hat F}_t$ and $\mathbf{\hat F}_h$.
As depicted in Fig.~\ref{method2}\textbf{(a)}, using $\mathbf{\hat F}_t$ and $\mathbf{\hat F}_h$ as inputs and initializing with $\mathbf{F}_i^0 = 0$, $\mathbf{z}_h^0 = 0$, and $\mathbf{z}_t^0 = 0$, in the first layer, $\mathbf{z}_t^0$ and $\mathbf{z}_h^0$ are fused with $\mathbf{\hat F}_t$ and $\mathbf{\hat F}_h$, respectively, resulting in $\mathbf{z}_t^1$ and $\mathbf{z}_h^1$. 
$\mathbf{z}_t^1$ and $\mathbf{z}_h^1$ are each multiplied by $\mathbf{F}_i^0$ and subsequently summed. The resultant feature is then fused with $\mathbf{\hat F}_t$ and $\mathbf{\hat F}_h$, yielding the manipulation feature $\mathbf{F}_i^1$ for the next layer.
This process is iteratively repeated, which can be formulated as:
\begin{equation}
    \mathbf{z}^0_t = 0, \quad \mathbf{z}^{k+1}_t = f_\theta (\mathbf{z}^k_t, \mathbf{\hat F}_t), \quad \mathbf{z}^*_t = f_\theta (\mathbf{z}^*_t, \mathbf{\hat F}_t),
\end{equation}
\begin{equation}
    \mathbf{z}^0_h = 0, \quad \mathbf{z}^{k+1}_h = f_\theta (\mathbf{z}^k_h, \mathbf{\hat F}_h), \quad \mathbf{z}^{*}_h = f_\theta (\mathbf{z}^*_h, \mathbf{\hat F}_h),
\end{equation}
\begin{equation}
    \mathbf{F}_i^0 = 0, \quad \mathbf{F}_i^{k+1} = f_\phi(\mathbf{\hat F}_t, \mathbf{\hat F}_h, f_\psi (\mathbf{F}_i^k, \mathbf{z}_t^{k+1}, \mathbf{z}_h^{k+1})), \quad \mathbf{F}_i^* = f_\phi(\mathbf{\hat F}_t, \mathbf{\hat F}_h, f_\psi (\mathbf{F}_i^*, \mathbf{z}_t^{*}, \mathbf{z}_h^{*})), 
\end{equation}
where $f_\psi$ represents the fusion unit which consists of element-wised multiplication and summation, while $f_\phi$ denotes nonlinear functions.
By utilizing the root-finding method described in Eq.~\ref{eq11}, the module ultimately derives the deeply fused feature $\mathbf{F}_i = \mathbf{F}_i^* $.
At this stage, the network has established a unidirectional constraint from interaction intention to manipulation. To create a more comprehensive constraint relationship between interaction elements and build a more stable joint prediction framework, we introduce residual connection units based on cross-attention blocks between $\mathbf{F}_i$ and $\mathbf{F}_t$, as well as between $\mathbf{F}_i$ and $\mathbf{F}_h$: 
\begin{equation}
    \overline{\mathbf{F}}_t = \mathbf{F}_t + \mathrm{MCA}(\mathbf{F}_t, \mathbf{F}_i),
    \quad
    \overline{\mathbf{F}}_h = \mathbf{F}_h + \mathrm{MCA}(\mathbf{F}_h, \mathbf{F}_i).
\end{equation}
Subsequently, we employ self-attention blocks and linear projection layers to decompose $\mathbf{F}_i$ into specific manipulation element features resulting in the extraction of hand pose feature $\mathbf{F}_p$, hand contact feature $\mathbf{F}_c$, and manipulation trajectory feature $\mathbf{F}_m$.
\subsection{Probabilistic Modeling Prediction Module}
\begin{figure}
    \centering
    \includegraphics[width=1\linewidth]{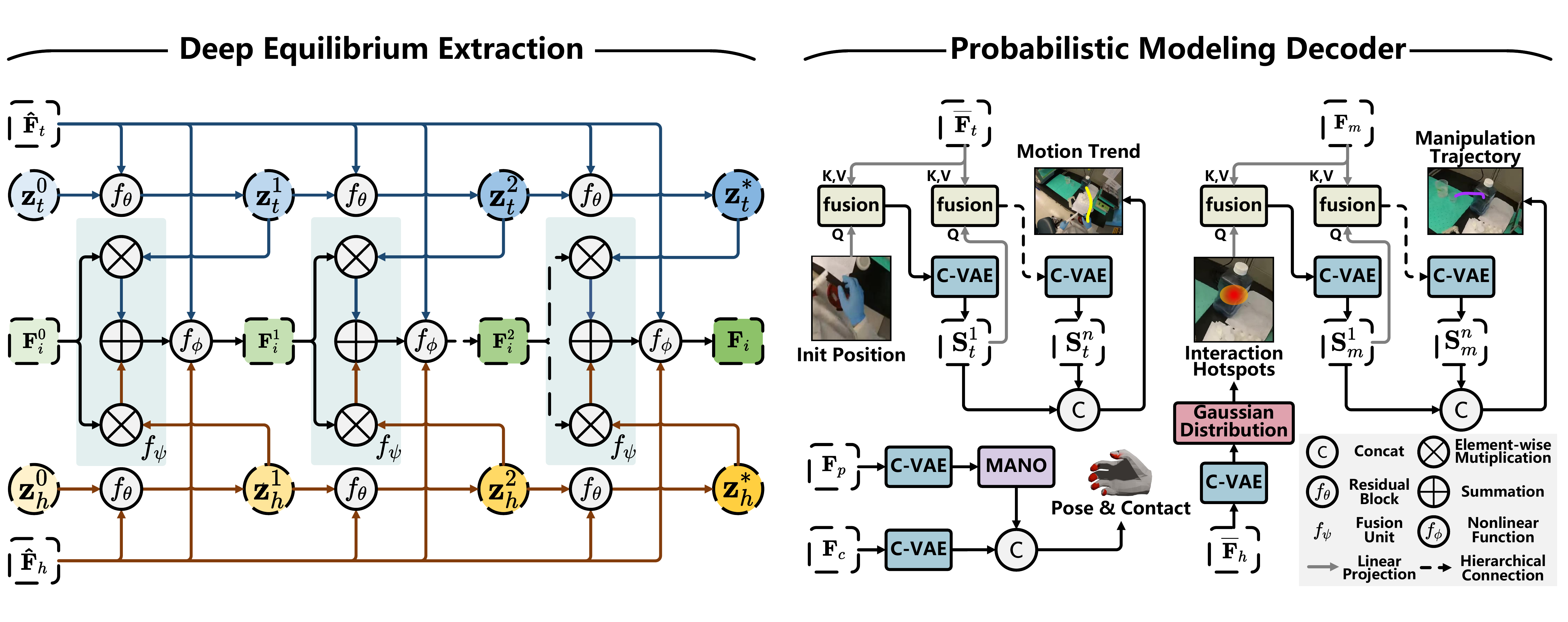}
    \put(-300,2){\makebox(0,0){\textbf{(a)}}}
    \put(-99,2){\makebox(0,0){\textbf{(b)}}}
    \caption{\textbf{Details of PEAR's structure}. \textbf{(a) DEQ Extraction model} is a crucial part of the Interaction Manipulation Extraction Module. This model takes the hand motion feature and the interaction hotspots feature as inputs, producing a fused feature in the equilibrium state, which functions as the manipulation feature. \textbf{(b) In the Probabilistic Modeling Prediction Module}, both hand motion trends and manipulation trajectories utilize chain-structured C-VAEs. The hand pose decoder incorporates the parametric MANO model to enhance the robustness of predictions.}
    \label{method2} 
\end{figure}
\label{sec3.4}
To capture the inherent randomness in human-object interaction processes, we employ C-VAEs~\cite{cvae} as the core of the anticipation decoder (Fig.~\ref{method2}(\textbf{b})).
A standard C-VAE follows an encoder-decoder structure.
Initially, the input $\mathbf{\Gamma}$ and condition $\mathbf{\Lambda}$ are passed through an encoding function $f_{e}$ to produce a normal distribution $\mathcal{N}$ with mean $\mu$ and co-variance $\sigma$, from which the latent variable $\mathbf{\Theta}$ is sampled. Subsequently, $\mathbf{\Theta}$ and $\mathbf{\Lambda}$ are processed through a decoding function $f_{d}$ to obtain the reconstructed output $\hat{\mathbf{\Gamma}}$.
\begin{equation}
    \mu,\sigma = f_{{e} }({\mathbf{\Gamma} },\mathbf{\Lambda}) , \quad\mathbf{\Theta }   \sim \mathcal{N}(\mu,\sigma),\quad \hat{\mathbf{\Gamma}} = f_{{d} }(\mathbf{\Theta }  ,\mathbf{\Lambda}).
    \label{eq17}
\end{equation}
For interaction hotspots, we utilize the refined $\overline {\mathbf{F}} _h$ as the condition to reconstruct the ground truth probability distribution and select the coordinates of the maximum value as the predicted contact point. 
Subsequently, centering on the contact point, we apply a Gaussian distribution to derive the interaction hotspots $\mathcal{O}$. 
For the hand motion trend, to ensure the continuity of the trajectory, we utilize sequential C-VAEs as the decoder. By conditioning on the corrected $\overline {\mathbf{F}} _t$ and the previous hand position, we iteratively predict the hand coordinates at the next keyframe, thereby generating the complete motion trajectory:
\begin{align}
    \mathbf{\Lambda } _t^i = \mathrm{MCA} (\mathbf{S} _t^{i-1},\overline{\mathbf{F}} _t), \quad \mathbf{S} _t^i = f _{d}(\mathbf{\Theta } ,\mathbf{\Lambda } _t^i), \quad  \mathcal{T} =\left \{ \mathbf S _t^0,\mathbf S _t^1,\cdots ,\mathbf S _t^{n_c} \right \},
\end{align}
where $\mathbf{S} _t^i$ denotes the hand location at the predicted time step $i$, $\mathbf{S} _t^0$ is the initial hand position in the given image $\mathit{I}$, and $n_c$ represents the contact time step.
For the manipulation trajectory $\mathcal{M}$, we utilize chain-structured C-VAEs akin to that employed for predicting hand motion trends. By conditioning on  $\mathbf{F} _m$ and utilizing the predicted contact point $ \mathbf{S} _t^{n_c}$ as the initial hand position, we iteratively generate the entire trajectory $\mathcal{M} = \left \{ \mathbf S _m^1,\mathbf S _m^2,\cdots ,\mathbf S _m^{n_m} \right \}$, where $\mathbf{S} _m^i$ represents the hand position at the anticipated time step $i$, and $n_m$ denotes the final predicted step of $\mathcal{M}$.
For 3D hand pose prediction, to better address challenges such as occlusion and blur, we introduce the parameterized hand model MANO~\cite{mano}, which provides a robust structure and extensive prior knowledge for hand pose prediction. 
Based on $ \mathbf{F} _p$, PEAR forecasts the pose parameters $\theta$ and shape parameters $\beta$ of the MANO model, thus anticipating the 3D hand pose $\mathcal{P}$.
For hand contact, using $ \mathbf{F} _c$ as the condition, we perform binary classification on the 778 vertex attributes of the MANO mesh to identify the contact regions $\mathcal{C}$.

\subsection{Loss Functions}
\label{sec3.5}
To enhance the overall accuracy of the joint prediction, the total training loss is formulated as follows:
\begin{equation}
    \mathcal{L } = \omega_1\mathcal{L} _t+ \omega_2\mathcal{L} _h + \omega_3\mathcal{L} _p + \omega_4\mathcal{L} _c + \omega_5\mathcal{L} _m,
\end{equation}
where $\mathcal{L}_t$, $\mathcal{L}_h$, $\mathcal{L}_p$, $\mathcal{L}_c$, and $\mathcal{L}_m$ represent the losses for hand motion trends $\mathcal{T}$, interaction hotspots $\mathcal{O}$, hand poses $\mathcal{P}$ with contact $\mathcal{C}$, and the manipulation trajectory $\mathcal{M}$, respectively. $\omega_{1-5}$ are hyper-parameters to balance the losses.
Specifically, since we employ C-VAEs as the backbone of the decoder, the loss for each interaction element during training includes a VAE-based loss, which comprises a reconstruction loss $\mathcal{L}_{recon}$ and a KL divergence $\mathcal{L}_{kl}$, as described by the following formula:
\begin{equation}
     \mathcal{L}_{kl}(\mu,\sigma)  =-\mathrm{KL} \left [ \mathcal{N}(\mu,\sigma)^{2}\left |  \right | \mathcal{N}(0,1)   \right ], \quad
      \mathcal{L}_{vae} =  \mathcal{L}_{recon} +\lambda  \mathcal{L}_{kl}.
\end{equation}
For interaction hotspots, hand motion trends, and manipulation trajectories, $\mathcal{L}_{recon}$ follows MSE loss:
\begin{equation}
    \mathcal{L}_{recon}  =|| \mathbf{\Gamma} -\hat{\mathbf{\Gamma} } || _2,
\end{equation}
where ${\mathbf{\Gamma}}$ denotes the ground truth and $\hat{\mathbf{\Gamma}}$ represents the predicted result.
For 3D hand poses, in addition to the MANO parameter reconstruction loss, which is calculated using the MSE loss, the total loss also includes a joint loss for the coordinates of 21 key points in the hand pose, which can be expressed as:
\begin{equation}
    \mathcal{L} _p =\mathcal{L} _{vae} +\zeta \mathcal{L} _{joint}, \quad
     \mathcal{L}_{joint}  =|| \mathbf{U} _{pred } - \mathbf{U} _{gt} || _1,
\end{equation}
where $\mathbf{U} _{pred }$ and $\mathbf{U} _{gt}$ are predicted 3D hand joint coordinates obtained through the MANO model and ground truth 3D hand locations, respectively.
For the hand contact, since the final output is a binary prediction, we introduce the BCE loss for the reconstruction loss function, following the formula below:
\begin{equation}
   \mathcal{L} _{recon} = -\frac{1}{n_o} \sum_{i=1}^{n_o} \left[ \mathbf{\rho}  _i \log(\hat{\mathbf{\rho} }_i) + (1 - \mathbf{\rho} _i) \log(1 - \hat{\mathbf{\rho} }_i) \right],
\end{equation}
where $n_o$ represents the number of vertices in the MANO mesh, $\mathbf{\rho}  _i$ denotes the ground truth contact probability, and $\hat{\mathbf{\rho}  _i}$ is the predicted probability that vertex is in a contact state.

\section{Datasets}
\subsection{Dataset Collection}
Based on hand-object interaction videos from EGO4D~\cite{ego4d}, we establish a comprehensive dataset, \textbf{EGO-HOIP}, for interaction anticipation. 
For each complete interaction process, we use the frame at the moment of hand-object contact as the reference point.
An image from 0.5 seconds before this reference frame is selected to represent the interaction scene, ensuring the interactive object is visible within the current field of view and providing a clear context for the interaction.
For each sample, we summarize the interaction information with a phrase consisting of a verb and a noun. 
The verb captures the human's potential action plan, while the noun denotes the potential interactive object. 
These phrases offer concise interaction prompts without introducing extraneous information. 
In total, we collect over 5,000 scene-prompt pairs to form the \textbf{EGO-HOIP} dataset, which enables detailed analysis and validation of our proposed model in anticipating hand-object interaction. 
\begin{figure}[t]
    \centering
    \includegraphics[width=0.9\linewidth]{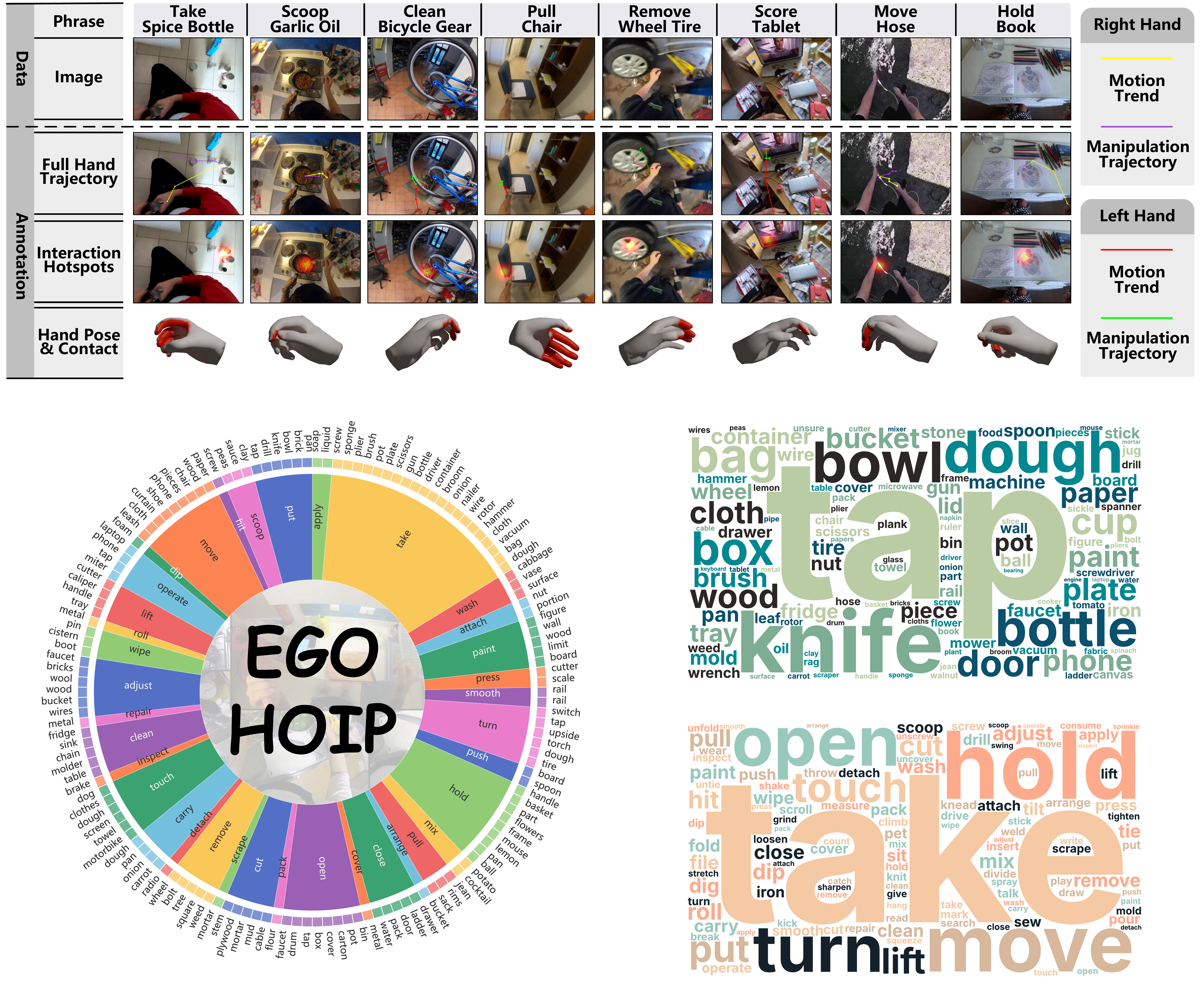}
    \put(-179.5,177){\makebox(0,0){\textbf{(a)}}}
    \put(-265,-2){\makebox(0,0){\textbf{(b)}}}
    \put(-87,89){\makebox(0,0){\textbf{(c)}}}
    \put(-87,-2){\makebox(0,0){\textbf{(d)}}}
    \caption{\textbf{EGO-HOIP Dataset}. \textbf{(a)} We detail the data and annotations in our dataset. Each sample comprises an interaction prompt phrase and a pre-interaction image representing the interaction scenario. We automatically generate annotations for hand trajectories both before and after contact, as well as for 3D hand poses. Additionally, we manually annotate interaction hotspots and the hand contact. \textbf{(b)} shows examples of verb-noun pairings in the phrases. \textbf{(c)} and \textbf{(d)} are word clouds for nouns and verbs, respectively, demonstrating the diversity of interaction prompts.}
    \label{fig:datasets}
\end{figure}
\subsection{Dataset Annotation}
We produce over 15k annotations of various types based on the collected data:
\textbf{1) Interaction Hotspots}: Due to the significant motion in the first-person video perspective, automated methods for annotating interaction hotspots result in cumulative errors. Therefore, we opt for manual annotation. Based on pre-contact interaction scenes, we manually mark the contact points on the objects and then apply a Gaussian distribution to these contact points to collect interaction hotspots annotations.
\textbf{2) Hand Motion Trends and Manipulation Trajectories}: For the hand trajectories in both pre-contact and post-contact phases, we employ an automated annotation method~\cite{both2}, marking three evenly spaced trajectory points for each phase to create a complete trajectory.
\textbf{3) 3D Hand Pose}: Given the occlusions and challenges in first-person images, we use the MANO~\cite{mano} parameterized model for hand pose estimation, HaMeR~\cite{hamer}, to automatically annotate  3D hand poses in contact frames.
\textbf{4) Hand Contact}: Similar to methods in~\cite{lemon, contactanno1}, hand contact annotations are based on expert knowledge. For each hand-object contact image, we provide the 3D hand mesh obtained from the previous step and the semantic interaction description. Professional annotators are hired to mark the vertices on the hand template that are in contact with objects. A secondary review of all hand contact annotations is conducted to ensure accuracy and consistency in annotation style.
\subsection{Statistic Analysis}
Fig.~\ref{fig:datasets}\textbf{(a)} shows a portion of the raw data along with various corresponding annotations, including indoor and outdoor scenes and examples of both left and right hands, demonstrating the extensive data breadth and annotation complexity of EGO-HOIP. Fig.~\ref{fig:datasets}\textbf{(b)} illustrates the combinations of various nouns and verbs in phrases, highlighting the high complexity of the phrases and the increased difficulty in anticipating hand-object interactions within this dataset. Fig.~\ref{fig:datasets}\textbf{(c)} and Fig.~\ref{fig:datasets}\textbf{(d)} display word clouds for nouns and verbs, respectively. The nouns exceed 600 in variety, while the verb categories surpass 90, indicating that similar phrases appear infrequently in EGO-HOIP. This feature allows for a better evaluation of a model’s generalization and robustness.

\section{Experiment}
\subsection{Benchmark Setting}
Our model is implemented in PyTorch and trained using the AdamW optimizer. The training process utilizes 4 NVIDIA 4090 GPUs for 75 epochs, with an initial learning rate set to $1e^{-4}$. Within the network architecture, we set $\mathbf{D} = 2$. The hyper-parameters in the loss function are configured as follows: \( w_{1-5} = 1 \), \( \lambda = 0.005 \), and \( \zeta = 1 \).
Furthermore, we select specific methods tailored to different interaction elements to facilitate comparative analysis. For interaction intention, we utilize \textbf{Hotspots}~\cite{aff1} for the isolated prediction of interaction hotspots and \textbf{USST}~\cite{3dtraj} for the isolated forecasting of hand motion trends. Additionally, we employ \textbf{FHOI}~\cite{both1}, \textbf{OCT}~\cite{both2}, and \textbf{BOT}~\cite{zzc} for the joint prediction of interaction intention. For interaction manipulation, we incorporate \textbf{I2L-MeshNet}~\cite{i2l}, \textbf{METRO}~\cite{METRO}, \textbf{MobRecon}~\cite{mobrecon}, and \textbf{HaMeR}~\cite{hamer} to predict 3D hand poses with contact. To forecast manipulation trajectories, we extend the hand motion trends generated by \textbf{USST}, \textbf{FHOI}, \textbf{OCT}, and \textbf{BOT} as comparative methods. Moreover, we implement image-text alignment and segmentation methods, including \textbf{CLIP}~\cite{clip}, \textbf{BLIP-2}~\cite{blip-2}, \textbf{CLIPSeg}~\cite{clipseg}, and \textbf{LISA}~\cite{lisa}, to jointly anticipate all interaction elements for both pre-contact and post-contact phases. To ensure a fair comparison, we integrate trainable decoders within these methods and apply balanced loss functions across the various elements.
\subsection{Evaluation Metric}
To present a comprehensive evaluation, we employ relevant metrics for different interaction elements. For hand motion trends and manipulation trajectories, we utilize Average Displacement Error (\textbf{ADE})~\cite{both2} and Final Displacement Error (\textbf{FDE})~\cite{both2} to measure overall and endpoint accuracy, respectively:
\begin{equation}
    \mathrm{ADE} = \frac{1}{n} \sum_{i=1}^{n} || \mathbf{p} _t - \hat{\mathbf{p} }_t ||_2, \quad
    \mathrm{FDE} = || \mathbf{p} _n - \hat{\mathbf{p} }_n ||_2,
\end{equation}
where $n$ denotes the total number of time steps, $\mathbf{p} _t$ represents the ground truth position, $\hat{\mathbf{p}}_t$ signifies the predicted position, ${\mathbf{p} _n}$ and $\hat{\mathbf{p}}_\mathrm{n}$ correspond the ground truth and predicted positions at time step $\mathrm{n}$, respectively.
Lower ADE and FDE values indicate higher trajectory accuracy.
For interaction hotspots, we employ Similarity (\textbf{SIM})~\cite{sim}, Area Under Curve-Judd (\textbf{AUC-J})~\cite{auc}, and Normalized Scanpath Saliency (\textbf{NSS})~\cite{nss}. 
SIM evaluates the overlap between predicted and ground truth distributions, AUC-J measures the proportion of true positive hotspots, and NSS assesses the alignment of predicted hotspots with human attention patterns. Higher SIM, AUC-J, and NSS values indicate better alignment between predicted and actual interaction hotspots.
For 3D hand poses, due to the ground truth being obtained in the contact frame and not aligned with the current view's camera coordinate system, we select Procrustes Aligned Mean Per Joint Position Error (\textbf{PA-MPJPE})~\cite{hamer} as the evaluation metric, defined as follows: 
\begin{equation}
    \text{PA-MPJPE} = \frac{1}{n_p} \sum_{i=1}^{n_p} || \mathbf{R} (\mathbf{p} _i) + \mathbf{v}  - \hat{\mathbf{p} }_i ||_2,
\end{equation}
where ${n_p}$ denotes the number of pose joints, $\mathbf{R}$ is the rotation matrix, $\mathbf{v}$ is the translation vector, $\mathbf{p} _i$ is the ground truth joint position, and $\hat{\mathbf{p} }_i$ is the predicted joint position. A smaller PA-MPJPE indicates a greater structural similarity between predicted and ground truth poses.
For hand contact, we employ \textbf{Precision}, \textbf{Recall}, and \textbf{F1 Score} as metrics. Precision reflects the proportion of correctly predicted contact regions out of all predicted contacts, Recall measures the proportion of actual contact regions correctly predicted, and F1 Score is the harmonic mean of Precision and Recall. Higher values in these metrics indicate better performance in predicting hand contact.
For different models, we repeat the inference 10 times and report the average results for a fair comparison.
\subsection{Experimental Results}
\begin{table}[h]
    \centering
    \caption{\textbf{The results of interaction intention anticipation}. We select relevant methods based on either video or image and text input. The best results are highlighted in \textbf{bold}, while the second-best results are \underline{underlined}. $\textcolor{darkpink}{\diamond}$ indicates the improvement relative to the first row. Our method achieves the best results in most metrics.}
    \renewcommand{\arraystretch}{1.4} 
    \setlength{\arrayrulewidth}{1.2pt} 
    \resizebox{\textwidth}{!}{%
    \begin{tabular}{cccc|ccccc}
    \toprule[1.5pt]
    \multicolumn{4}{c|}{\textbf{Hand Motion Trend}} & \multicolumn{5}{c}{\textbf{Interaction Hotspots}} \\ 
    \midrule[1.2pt]
    \textbf{Methods} & \textbf{Modality} & \textbf{ADE} $\downarrow$ & \textbf{FDE} $\downarrow$ & \textbf{Methods} & \textbf{Modality} & \textbf{SIM} $\uparrow$ & \textbf{AUC-J} $\uparrow$ & \textbf{NSS} $\uparrow$ \\ 
    \midrule[1.2pt]
    FHOI\cite{both1} & Video & 0.182 & 0.180 
     & Hotspots\cite{aff1} & Video & 0.163 & 0.657 & 0.572 \\ 
    USST\cite{3dtraj} & Video & $0.138\textcolor{darkpink}{\scriptstyle~\diamond24.2\%}$ & $0.154\textcolor{darkpink}{\scriptstyle~\diamond14.4\%}$ & FHOI\cite{both1} & Video & $0.192\textcolor{darkpink}{\scriptstyle~\diamond17.8\%}$ & $0.685\textcolor{darkpink}{\scriptstyle~\diamond4.3\%}$ & $0.872\textcolor{darkpink}{\scriptstyle~\diamond52.4\%}$ \\ 
    OCT\cite{both2} & Video & $0.134\textcolor{darkpink}{\scriptstyle~\diamond26.4\%}$ & $0.147\textcolor{darkpink}{\scriptstyle~\diamond18.3\%}$ & OCT\cite{both2}  & Video & $0.278\textcolor{darkpink}{\scriptstyle~\diamond70.6\%}$ & $0.749\textcolor{darkpink}{\scriptstyle~\diamond14.0\%}$ & $1.418\textcolor{darkpink}{\scriptstyle~\diamond147.9\%}$ \\ 
    BOT\cite{zzc} & Video & $\textbf{0.118}\textcolor{darkpink}{\scriptstyle~\diamond35.2\%}$ & $0.137\textcolor{darkpink}{\scriptstyle~\diamond23.9\%}$ & BOT\cite{zzc} & Video &$0.304\textcolor{darkpink}{\scriptstyle~\diamond86.5\%}$ & $0.763\textcolor{darkpink}{\scriptstyle~\diamond16.1\%}$ & $1.614\textcolor{darkpink}{\scriptstyle~\diamond182.2\%}$ \\ 
    CLIP\cite{clip} & Image+Text & $0.163\textcolor{darkpink}{\scriptstyle~\diamond10.4\%}$ & $0.134\textcolor{darkpink}{\scriptstyle~\diamond25.6\%}$ & CLIP\cite{clip} & Image+Text & $0.325\textcolor{darkpink}{\scriptstyle~\diamond99.4\%}$ & $0.795\textcolor{darkpink}{\scriptstyle~\diamond21.0\%}$ & $1.746\textcolor{darkpink}{\scriptstyle~\diamond205.2\%}$ \\ 
    BLIP-2\cite{blip-2} & Image+Text & $0.162\textcolor{darkpink}{\scriptstyle~\diamond11.0\%}$ & $0.132\textcolor{darkpink}{\scriptstyle~\diamond26.7\%}$ & BLIP-2\cite{blip-2} & Image+Text & $0.332\textcolor{darkpink}{\scriptstyle~\diamond103.7\%}$ & $0.802\textcolor{darkpink}{\scriptstyle~\diamond22.1\%}$ & $1.821\textcolor{darkpink}{\scriptstyle~\diamond218.4\%}$ \\ 
    CLIPSeg\cite{clipseg} & Image+Text & $0.154\textcolor{darkpink}{\scriptstyle~\diamond15.4\%}$ & $0.128\textcolor{darkpink}{\scriptstyle~\diamond28.9\%}$ & CLIPSeg\cite{clipseg} & Image+Text & $0.362\textcolor{darkpink}{\scriptstyle~\diamond122.1\%}$ & $0.829\textcolor{darkpink}{\scriptstyle~\diamond26.2\%}$ & $2.084\textcolor{darkpink}{\scriptstyle~\diamond264.3\%}$ \\ 
    LISA\cite{lisa} & Image+Text & $0.150\textcolor{darkpink}{\scriptstyle~\diamond19.2\%}$ & $\underline{0.122}\textcolor{darkpink}{\scriptstyle~\diamond32.2\%}$ & LISA\cite{lisa} & Image+Text & $\underline{0.389}\textcolor{darkpink}{\scriptstyle~\diamond138.7\%}$ & $\underline{0.843}\textcolor{darkpink}{\scriptstyle~\diamond28.2\%}$ & $\underline{2.291}\textcolor{darkpink}{\scriptstyle~\diamond300.5\%}$ \\ 
    \rowcolor{lightgray}Ours & Image+Text & $\underline{0.126}\textcolor{darkpink}{\scriptstyle~\diamond30.8\%}$ & $\textbf{0.104}\textcolor{darkpink}{\scriptstyle~\diamond42.2\%}$ & Ours & Image+Text & $\textbf{0.449}\textcolor{darkpink}{\scriptstyle~\diamond171.8\%}$ & $\textbf{0.877}\textcolor{darkpink}{\scriptstyle~\diamond34.4\%}$ &$\textbf{2.726}\textcolor{darkpink}{\scriptstyle~\diamond377.1\%}$ \\ 
    \bottomrule[1.5pt]
    \end{tabular}
    }
    \label{tab1}
\end{table}
We present the prediction results for pre-contact interaction elements using different methods in Tab.~\ref{tab1}. Our method achieves SOTA performance across most metrics, except for the ADE in hand motion trend prediction, where video-based methods have a slight advantage. Compared to video-based methods, since the first half of the hand motion trend is mainly influenced by the hand's motion inertia, video-based prediction methods leverage a past sequence of hand movements, incorporating this inertia into the interaction scenario, thus achieving higher short-term trajectory prediction accuracy and an advantage in ADE. However, our method excels in the FDE for hand motion trend and all metrics for interaction hotspots due to effectively utilizing the noun in the phrase to guide the interactive object, thereby refining the trajectory's endpoint and improving the prediction accuracy of the contact area. 
Moreover, compared to image-text matching methods, our approach offers significant advantages. This is due to our strategy of decomposing the phrase into verbs and nouns and performing cross-alignment of verbs, nouns, and images within the Interaction Intention Excavation Module. This process effectively overcomes intention uncertainty and enhances the accuracy of predicting hand motion trends and interaction hotspots.
\begin{table}[h]
    \centering
    \caption{\textbf{The results of interaction manipulation anticipation}. We choose relevant methods based on images, videos, and text data. The best results are highlighted in \textbf{bold}, while the second-best results are \underline{underlined}. $\textcolor{darkpink}{\diamond}$ indicates the improvement relative to the first row. ``Img.'' denotes image, while ``T.'' signifies text. \textbf{PEAR} achieves the best results across all metrics.}
    \renewcommand{\arraystretch}{1.5} 
    \setlength{\arrayrulewidth}{1.2pt} 
    \resizebox{\textwidth}{!}{%
    \begin{tabular}{cccccc|cccc}
    \toprule[1.5pt]
    \multicolumn{6}{c|}{\textbf{Hand Pose With Contact}} & \multicolumn{4}{c}{\textbf{Manipulation Trajectory}} \\ 
    \midrule[1.2pt]
    \textbf{Methods} & \textbf{Modality} & \textbf{PA-MPJPE} $\downarrow$ & \textbf{Precision} $\uparrow$ & \textbf{Recall} $\uparrow$ & \textbf{F1} $\uparrow$ & \textbf{Methods} & \textbf{Modality} & \textbf{ADE} $\downarrow$ & \textbf{FDE} $\downarrow$ \\ 
    \midrule[1.2pt]
    I2L-MeshNet\cite{i2l} & Image  & $13.363$ & $0.738$ & $0.729$ & $0.674$ & FHOI~\cite{both1} & Video & $0.204$ & $0.212$  \\ 
    METRO\cite{METRO} & Image & $13.131\textcolor{darkpink}{\scriptstyle~\diamond1.7\%}$ & $0.754\textcolor{darkpink}{\scriptstyle~\diamond2.2\%}$ & $0.735\textcolor{darkpink}{\scriptstyle~\diamond0.8\%}$ & $0.686\textcolor{darkpink}{\scriptstyle~\diamond1.8\%}$ & USST\cite{3dtraj} & Video & $0.177\textcolor{darkpink}{\scriptstyle~\diamond13.2\%}$ & $0.180\textcolor{darkpink}{\scriptstyle~\diamond15.1\%}$ \\ 
    MobRecon\cite{mobrecon} & Image & $12.783\textcolor{darkpink}{\scriptstyle~\diamond4.3\%}$ & $0.744\textcolor{darkpink}{\scriptstyle~\diamond0.8\%}$ & $0.748\textcolor{darkpink}{\scriptstyle~\diamond2.6\%}$ & $0.688\textcolor{darkpink}{\scriptstyle~\diamond2.1\%}$ & OCT\cite{both2} & Video & $0.172\textcolor{darkpink}{\scriptstyle~\diamond15.7\%}$ & $0.174\textcolor{darkpink}{\scriptstyle~\diamond17.9\%}$ \\ 
    HaMeR\cite{hamer} & Image & $12.512\textcolor{darkpink}{\scriptstyle~\diamond6.4\%}$ & $0.756\textcolor{darkpink}{\scriptstyle~\diamond2.4\%}$ & $0.741\textcolor{darkpink}{\scriptstyle~\diamond1.6\%}$ & $0.691\textcolor{darkpink}{\scriptstyle~\diamond2.5\%}$ & BOT\cite{zzc} & Video & $0.167\textcolor{darkpink}{\scriptstyle~\diamond18.1\%}$ & $0.169\textcolor{darkpink}{\scriptstyle~\diamond20.3\%}$ \\ 
    CLIP\cite{clip} & Img.+T. & $12.765\textcolor{darkpink}{\scriptstyle~\diamond4.5\%}$ & $0.752\textcolor{darkpink}{\scriptstyle~\diamond1.9\%}$ & $0.737\textcolor{darkpink}{\scriptstyle~\diamond1.1\%}$ & $0.685\textcolor{darkpink}{\scriptstyle~\diamond1.6\%}$ & CLIP\cite{clip} & Img.+T. & $0.182\textcolor{darkpink}{\scriptstyle~\diamond10.8\%}$ & $0.202\textcolor{darkpink}{\scriptstyle~\diamond4.7\%}$ \\ 
    BLIP-2\cite{blip-2} & Img.+T. & $12.749\textcolor{darkpink}{\scriptstyle~\diamond4.6\%}$ & $0.755\textcolor{darkpink}{\scriptstyle~\diamond2.3\%}$ & $0.742\textcolor{darkpink}{\scriptstyle~\diamond1.8\%}$ & $0.688\textcolor{darkpink}{\scriptstyle~\diamond2.1\%}$ & BLIP-2\cite{blip-2} & Img.+T. & $0.177\textcolor{darkpink}{\scriptstyle~\diamond13.2\%}$ & $0.197\textcolor{darkpink}{\scriptstyle~\diamond7.1\%}$ \\ 
    CLIPSeg\cite{clipseg} & Img.+T. & $12.326\textcolor{darkpink}{\scriptstyle~\diamond7.8\%}$ & $0.762\textcolor{darkpink}{\scriptstyle~\diamond3.3\%}$ & $0.748\textcolor{darkpink}{\scriptstyle~\diamond2.6\%}$ & $0.696\textcolor{darkpink}{\scriptstyle~\diamond3.3\%}$ & CLIPSeg\cite{clipseg} & Img.+T. & $0.165\textcolor{darkpink}{\scriptstyle~\diamond19.1\%}$ & $0.174\textcolor{darkpink}{\scriptstyle~\diamond17.9\%}$ \\ 
    LISA\cite{lisa} & Img.+T. & $\underline{12.132}\textcolor{darkpink}{\scriptstyle~\diamond9.2\%}$ & $\underline{0.774}\textcolor{darkpink}{\scriptstyle~\diamond4.9\%}$ & $\underline{0.752}\textcolor{darkpink}{\scriptstyle~\diamond3.2\%}$ & $\underline{0.704}\textcolor{darkpink}{\scriptstyle~\diamond4.5\%}$ & LISA\cite{lisa} & Img.+T. & $\underline{0.160}\textcolor{darkpink}{\scriptstyle~\diamond21.6\%}$ & $\underline{0.167}\textcolor{darkpink}{\scriptstyle~\diamond21.2\%}$ \\ 
    \rowcolor{lightgray} Ours & Img.+T. & $\textbf{10.757}\textcolor{darkpink}{\scriptstyle~\diamond16.1\%}$ & $\textbf{0.804}\textcolor{darkpink}{\scriptstyle~\diamond8.9\%}$ & $\textbf{0.765}\textcolor{darkpink}{\scriptstyle~\diamond4.9\%}$ & $\textbf{0.725}\textcolor{darkpink}{\scriptstyle~\diamond7.6\%}$ & Ours & Img.+T. & $\textbf{0.143}\textcolor{darkpink}{\scriptstyle~\diamond29.9\%}$ & $\textbf{0.148}\textcolor{darkpink}{\scriptstyle~\diamond30.2\%}$ \\ 
    \bottomrule[1.5pt]
    \end{tabular}
    }
    \label{tab2}
\end{table}

Tab.~\ref{tab2} presents the prediction results of interaction manipulation using different methods. Our method achieves SOTA performance across all metrics for these elements. 
Compared to image-based hand pose estimation methods, our approach demonstrates a significant advantage in predicting hand poses with contact. 
This superiority stems from the fact that image-based methods do not effectively capture interaction intent, resulting in imprecise localization of the interactive object and subsequently diminishing overall prediction accuracy.
In comparison to video-based prediction methods, our approach excels in predicting manipulation trajectories. This is attributed to our consideration of the bidirectional constraint relationship between interaction intention and manipulation, which effectively mitigates manipulation uncertainty. 
Conversely, video-based methods lack the necessary constraint modules, causing prediction errors to accumulate as the interaction process progresses, resulting in less accurate manipulation trajectory predictions compared to those achieved by PEAR.
Moreover, compared to other image-text alignment methods, our approach demonstrates a distinct advantage in predicting all elements of manipulation.
This advantage is attributed to our use of the DEQ extraction model, which extracts stable and precise manipulation features.
Furthermore, we establish bidirectional constraints between intention and manipulation utilizing residual connections, thereby reducing manipulation uncertainty and enhancing prediction accuracy.

\subsection{Visualization Results}
\begin{figure}[t]
    \centering
    \includegraphics[width=1\linewidth]{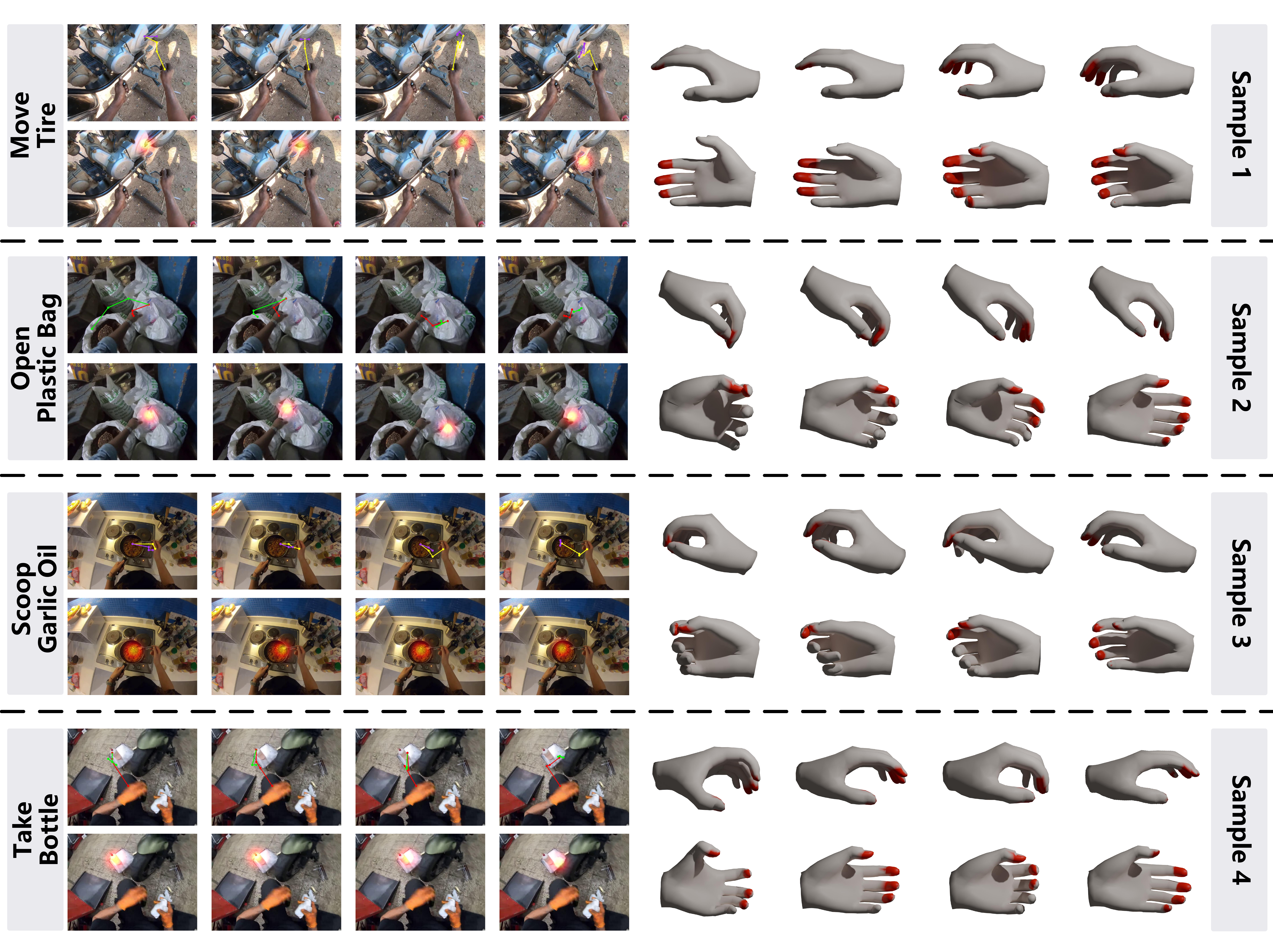}
    \put(-356,297){\makebox(0,0){\footnotesize\textbf{GT}}}
    \put(-312,297){\makebox(0,0){\footnotesize\textbf{Ours}}}
    \put(-266,297){\makebox(0,0){\footnotesize\textbf{LISA}\cite{lisa}}}
    \put(-221,297){\makebox(0,0){\footnotesize\textbf{BOT}\cite{zzc}}}
    \put(-177,297){\makebox(0,0){\footnotesize\textbf{GT}}}
    \put(-134,297){\makebox(0,0){\footnotesize\textbf{Ours}}}
    \put(-87.5,297){\makebox(0,0){\footnotesize\textbf{LISA}\cite{lisa}}}
    \put(-40.5,297){\makebox(0,0){\footnotesize\textbf{HaMeR}\cite{hamer}}}                                               
    \caption{\textbf{Visualization results.} We select four samples, labeled sequentially from top to bottom. In each sample, the left side depicts the predictions of the complete hand trajectory and interaction hotspots, while the right side presents the predicted hand poses with contact from two different perspectives. \textcolor{myyellow}{Yellow} and \textcolor{red}{red} denote the right and left-hand motion trends, respectively, while \textcolor{mypurple}{purple} and \textcolor{green}{green} represent the right and left manipulation trajectories, respectively. Additionally, the areas of contact with the object are marked in \textcolor{red}{red} on the hand mesh.}
    \label{viz1}
\end{figure}
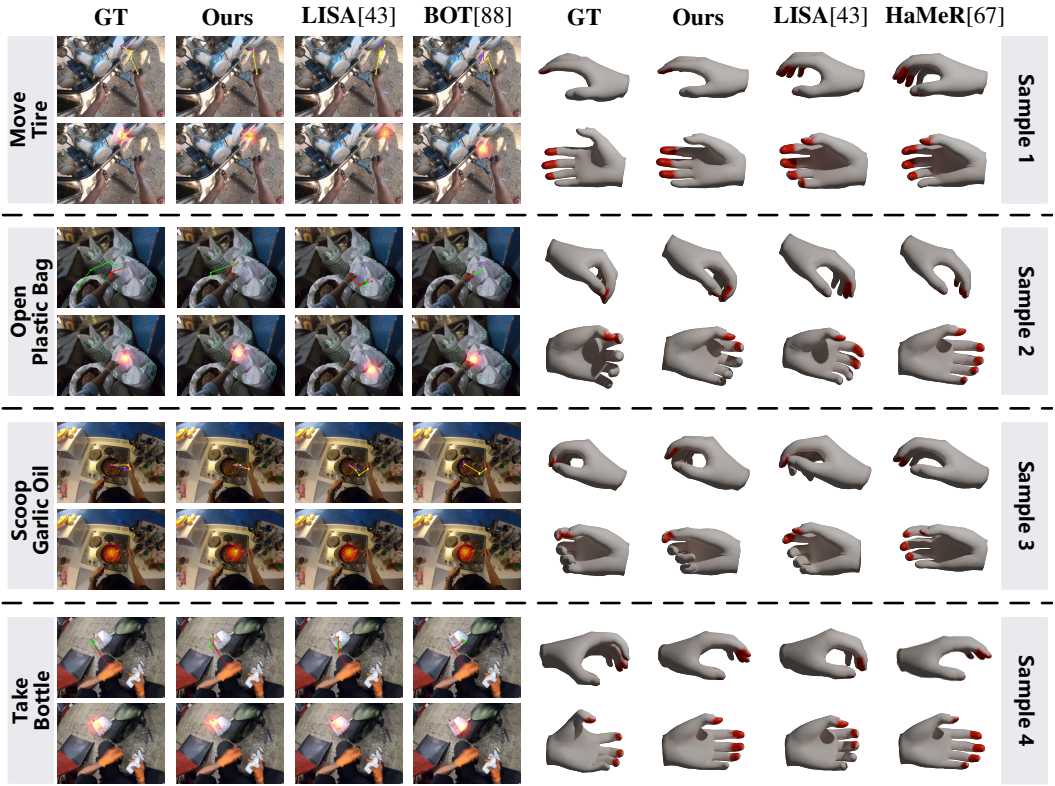
As illustrated in Fig.~\ref{viz1}, we select GT (Ground Truth), PEAR, and LISA (the second-best image-text prediction method) for comparative visualization of all interaction elements (columns 1, 2, 3, and 5, 6, 7, respectively). 
Additionally, we choose BOT for comparing complete trajectories and interaction hotspots (column 4) and HaMeR for comparing hand poses with contact (column 8). The results indicate that our method demonstrates a clear overall advantage.

Compared to BOT, PEAR demonstrates significant advantages in predicting interaction hotspots and manipulation trajectories.
BOT's lack of phrase utilization results in inaccuracies in identifying interactive objects, as observed in the first and fourth samples, thereby increasing errors in contact regions.
Additionally, BOT's trajectory predictions tend to follow a single direction due to the absence of guidance between intention and manipulation, leading to errors, as evidenced in the second sample.
In comparison to HaMeR, PEAR excels in predicting hand poses with contact. HaMeR's predictions, primarily based on the current scene, fail to effectively extract interaction intention, resulting in average hand poses and contacts, such as slightly bent fingers and constant thumb contact. 
In contrast, our method effectively leverages verbs from phrases to plan actions, resulting in more targeted and accurate hand poses with contact.
For example, in the first sample, the predicted hand pose is relaxed with no thumb contact, while in the second and third samples, the hand is pinching, with the thumb and index finger touching.

Compared to the image-text segmentation model LISA, our method exhibits superior performance in predicting all interaction elements, particularly when handling larger objects.
In the first sample, our method's prediction of interaction hotspots is more precise, resulting in a hand pose with contact that closely aligns with the ground truth. 
In the second sample, due to the extensive area of the plastic bag, although LISA accurately identifies the interactive object, it fails to analyze the object's functionality concerning the human's intention.
This shortcoming prevents LISA from narrowing the contact area, leading to significant deviations across all interaction elements.
In summary, PEAR not only leverages both image and phrase information but also introduces comprehensive and stable constraints between different interaction elements.
These constraints allow the elements to mutually correct each other, thereby improving overall prediction coherence and accuracy.

\section{Analytical Experiment}
\subsection{Ablation Study}
An ablation study is conducted on the Interaction Intention Excavation Module and the Interaction Manipulation Extraction Module within the pipeline to examine the units that establish constraints between different interaction elements. 
In the proposed method, three primary constraint relationships are established: the constraint between hand motion trends and interaction hotspots (parallel cross-attention blocks), the extraction of manipulation elements by interaction intention (DEQ extraction model), and the correction of intention elements by the manipulation (residual connection units).
Each of these modules is systematically removed, either individually or in pairs, and the experimental results are shown in Tab.~\ref{tab3}.
The results clearly show that the overall prediction performance is optimal when all three constraints are present.
Removing the direct constraint relationship between hand motion trends and interaction hotspots leads to a general decline in overall prediction accuracy. 
This finding indicates that aligning these two elements helps mitigate the variability of intention elements, thereby improving the prediction accuracy of pre-contact interaction intention and facilitating the subsequent prediction of manipulation elements.
The removal of the DEQ extraction model results in a significant decrease in the accuracy of the corresponding manipulation elements.
This reduction in accuracy is due to the lack of constraints from the contact area and hand motion trends, forcing the prediction to rely solely on image and phrase features, which lack detailed guidance.
Omitting the residual connection units leads to a substantial degradation in the prediction results for intention elements, underscoring the importance of this module in refining hand motion trends and interaction hotspots.

\begin{table}[h]
    \centering
    \caption{\textbf{Ablation study}. We conduct ablation experiments on the three main constraint chains in the pipeline by removing one or two of them to study their impact on network performance. ``\textbf{Cross}'' represents the constraints between intention elements, ``\textbf{DEQ.}'' signifies the module guiding manipulation by interaction intention, and ``\textbf{Res.}'' denotes the unit correcting intention by manipulation. ``\checkmark'' indicates that the module is included in the network. The best results are highlighted in \textbf{bold}.}
    \renewcommand{\arraystretch}{1.5} 
    \setlength{\arrayrulewidth}{1.2pt} 
    \resizebox{\textwidth}{!}{%
    \begin{tabular}{ccc|cc|ccc|cccc|cc}
    \toprule[1.5pt]
    \multicolumn{3}{c|}{\textbf{Constraints}} & \multicolumn{2}{c|}{\textbf{Motion Trend}} & \multicolumn{3}{c|}{\textbf{Interaction Hotspots}} & \multicolumn{4}{c|}{\textbf{Hand Pose With Contact}} & \multicolumn{2}{c}{\textbf{Mani Trajectory}} \\ 
    \midrule[1.2pt]
    \textbf{Cross.} & \textbf{DEQ.} & \textbf{Res.} & \textbf{ADE} $\downarrow$ & \textbf{FDE} $\downarrow$ & \textbf{SIM} $\uparrow$ & \textbf{AUC-J} $\uparrow$ & \textbf{NSS} $\uparrow$ & \textbf{PA.} $\downarrow$ & \textbf{Pre.} $\uparrow$ & \textbf{Rec.} $\uparrow$ & \textbf{F1} $\uparrow$ & \textbf{ADE} $\downarrow$ & \textbf{FDE} $\downarrow$ \\ 
    \midrule[1.2pt]
    \checkmark &  &  & 0.143 & 0.119 & 0.403 & 0.848 & 2.259 & 12.157 & 0.759 & 0.746 & 0.697 & 0.157 & 0.172 \\ 
     & \checkmark &  & 0.145 & 0.119 & 0.403 & 0.851 & 2.245 & 11.926 & 0.770 & 0.742 & 0.705 & 0.154 & 0.166 \\ 
     &  &  \checkmark& 0.145 & 0.121 & 0.398 & 0.849 & 2.213 & 12.283 & 0.763 & 0.740 & 0.694 & 0.156 & 0.171 \\
    \checkmark & \checkmark & & 0.136 & 0.114 & 0.427 & 0.872 & 2.614 & 11.639 & 0.781 & 0.755 & 0.713 & 0.152 & 0.159 \\
    \checkmark & & \checkmark & 0.138 & 0.115 & 0.423 & 0.869 & 2.573 & 11.972 & 0.767 & 0.748 & 0.701 & 0.152 & 0.166 \\
     & \checkmark & \checkmark & 0.142 & 0.118 & 0.408 & 0.855 & 2.432 & 11.879 & 0.779 & 0.753 & 0.711 & 0.151 & 0.163 \\
    \rowcolor{lightgray} \checkmark & \checkmark & \checkmark & \textbf{0.126} & \textbf{0.104} & \textbf{0.449} & \textbf{0.877} & \textbf{2.726} & \textbf{10.757} & \textbf{0.804} & \textbf{0.765} & \textbf{0.725} & \textbf{0.143} & \textbf{0.148} \\
    \bottomrule[1.5pt]
    \end{tabular}
    }
    \label{tab3}
\end{table}
\subsection{Modification Study}
Building upon the ablation study, further research investigates methods for establishing constraints between interaction elements by substituting specific units within these three modules, the results are presented in Tab.~\ref{tab4}. 
The methods compared include summation (\textbf{Sum.}), concatenation (\textbf{Concat.}), and series cross-attention blocks (\textbf{Cross-att.}), with the introduction of 1x1 convolutional kernels and MLPs to adjust the number of output features and increase the depth of the fusion units as needed.
Firstly, regarding the constraint between hand motion trends and interaction hotspots, the proposed method using parallel cross-attention blocks demonstrates a clear advantage. 
This is because these two elements represent projections of hand motion patterns and object functional attributes, respectively, creating a significant feature-level disparity.
Over-fusing these features can lead to interference and information loss. 
The proposed parallel cross-attention blocks introduce cross-features while preserving the original features, thus avoiding interference during constraint establishment and yielding superior prediction outcomes.
For the constraint from intention to manipulation, the proposed DEQ extraction model achieves optimal results. 
This is because the pre-contact features exhibit dynamic characteristics, and the ``infinite-depth'' network simulation can dynamically adjust the network depth when extracting manipulation element features, thereby solving for a stable feature in an equilibrium state. Fixed-depth networks often struggle to equally accommodate all input features during the extraction of fused features, resulting in inaccuracies and increased prediction errors.
For the constraint from manipulation to intention, the residual connection units produce the best results.
This indicates that when establishing this constraint relationship, the original interaction intention element features should remain dominant, as over-introduction of manipulation features can reduce overall prediction accuracy.
\begin{table}[h]
    \centering
    \caption{\textbf{Modification study}. We replace three main constraint modules to study the impact of different constraint methods on predictions. ``\textbf{Cross.}'' indicates the constraints between intention elements, ``\textbf{DEQ.}'' signifies the module guiding manipulation by intention, and ``\textbf{Res.}'' represents the unit correcting intention by manipulation. The best results are highlighted in \textbf{bold}.}
    \renewcommand{\arraystretch}{1.5} 
    \setlength{\arrayrulewidth}{1.2pt} 
    \setlength{\aboverulesep}{0pt} 
    \setlength{\belowrulesep}{0pt} 
    \resizebox{\textwidth}{!}{%
    \begin{tabular}{cc|cc|ccc|cccc|cc}
    \toprule[1.5pt]
    \multicolumn{2}{c|}{\textbf{Constraints}} & \multicolumn{2}{c|}{\textbf{Motion Trend}} & \multicolumn{3}{c|}{\textbf{Interaction Hotspots}} & \multicolumn{4}{c|}{\textbf{Hand Pose With Contact}} & \multicolumn{2}{c}{\textbf{Mani. Trajectory}} \\ 
   \midrule[1.2pt]
   \textbf{module} & \textbf{methods} & \textbf{ADE} $\downarrow$ & \textbf{FDE} $\downarrow$ & \textbf{SIM} $\uparrow$ & \textbf{AUC-J} $\uparrow$ & \textbf{NSS} $\uparrow$ & \textbf{PA.} $\downarrow$ & \textbf{Pre.} $\uparrow$ & \textbf{Rec.} $\uparrow$ & \textbf{F1} $\uparrow$ & \textbf{ADE} $\downarrow$ & \textbf{FDE} $\downarrow$ \\ 
   \midrule[1.2pt]
   \multirow{4}{*}{\textbf{Cross.}} & Sum. & 0.142 & 0.113 & 0.418 & 0.869 & 2.541 & 12.023 & 0.782 & 0.727 & 0.692 & 0.156 & 0.168 \\ 
    & Concat. & 0.138 & 0.110 & 0.423 & 0.865 & 2.555 & 11.782 & 0.778 & 0.736 & 0.694 & 0.154 & 0.159 \\ 
    & Cross-att. & 0.136 & 0.108 & 0.429 & 0.868 & 2.596 & 11.325 & 0.785 & 0.741 & 0.709 & 0.158 & 0.164 \\ 
   \rowcolor{lightgray} & Ours & \textbf{0.126} & \textbf{0.104} & \textbf{0.449} & \textbf{0.877} & \textbf{2.726} & \textbf{10.757} & \textbf{0.804} & \textbf{0.765} & \textbf{0.725} & \textbf{0.143} & \textbf{0.148} \\
    \midrule[1.2pt]
     \multirow{4}{*}{\textbf{DEQ.}} & Sum. & 0.144 & 0.116 & 0.422 & 0.866 & 2.543 & 11.449 & 0.773 & 0.738 & 0.694 & 0.158 & 0.161 \\ 
    & Concat. & 0.139 & 0.113 & 0.425 & 0.857 & 2.464 & 11.337 & 0.780 & 0.746 & 0.703 & 0.151 & 0.167 \\ 
    & Cross-att. & 0.142 & 0.115 & 0.423 & 0.865 & 2.561 & 11.259 & 0.784 & 0.749 & 0.711 & 0.149 & 0.162 \\ 
   \rowcolor{lightgray} & Ours & \textbf{0.126} & \textbf{0.104} & \textbf{0.449} & \textbf{0.877} & \textbf{2.726} & \textbf{10.757} & \textbf{0.804} & \textbf{0.765} & \textbf{0.725} & \textbf{0.143} & \textbf{0.148} \\
    \midrule[1.2pt]
    \multirow{4}{*}{\textbf{Res.}} & Sum. & 0.141 & 0.114 & 0.418 & 0.863 & 2.421 & 11.673 & 0.775 & 0.743 & 0.700 & 0.152 & 0.160 \\ 
    & Concat. & 0.138 & 0.113 & 0.414 & 0.866 & 2.416 & 11.215 & 0.780 & 0.754 & 0.707 & 0.153 & 0.164 \\ 
    & Cross-att. & 0.136 & 0.109 & 0.425 & 0.867 & 2.574 & 11.472 & 0.778 & 0.746 & 0.704 & 0.155 & 0.163 \\ 
  \rowcolor{lightgray}  & Ours & \textbf{0.126} & \textbf{0.104} & \textbf{0.449} & \textbf{0.877} & \textbf{2.726} & \textbf{10.757} & \textbf{0.804} & \textbf{0.765} & \textbf{0.725} & \textbf{0.143} & \textbf{0.148} \\
    \bottomrule[1.5pt]
    \end{tabular}
    }
    \label{tab4}
\end{table}

\subsection{Hyper-parameter Study}
\begin{figure}[h]
    \centering
    \includegraphics[width=1\linewidth]{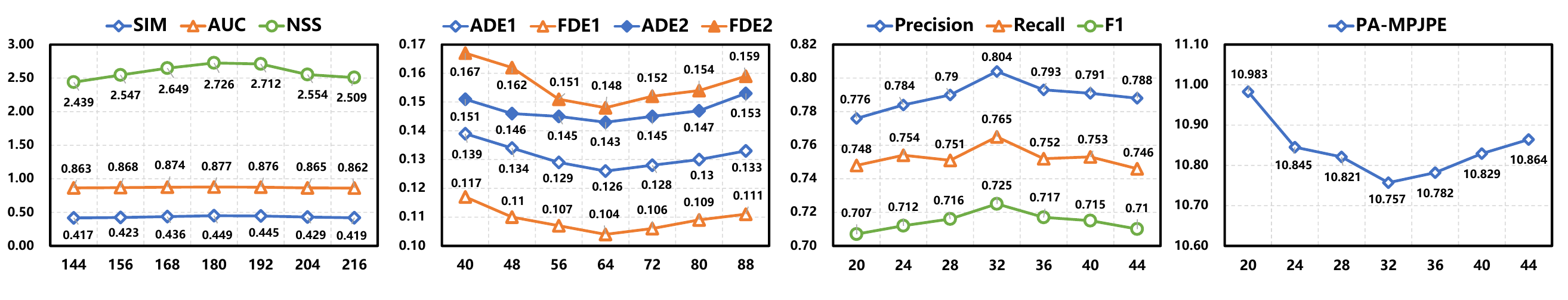}
    \put(-345,0){\makebox(0,0){\textbf{(a)}}}
    \put(-246,0){\makebox(0,0){\textbf{(b)}}}
    \put(-146,0){\makebox(0,0){\textbf{(c)}}}
    \put(-46,0){\makebox(0,0){\textbf{(d)}}}
    \caption{\textbf{(a)}, \textbf{(b)}, \textbf{(c)}, and \textbf{(d)} represent the experimental results for interaction hotspots, complete hand trajectories, hand contact, and hand poses, respectively. In each chart, the horizontal axis represents the latent dimensions, while the vertical axis indicates the values of metrics. ADE1 and FDE1 represent the results of pre-contact hand motion trends, respectively, while ADE2 and FDE2 stand for the results of post-contact manipulation trajectories, respectively.}
    \label{ablation}
\end{figure}
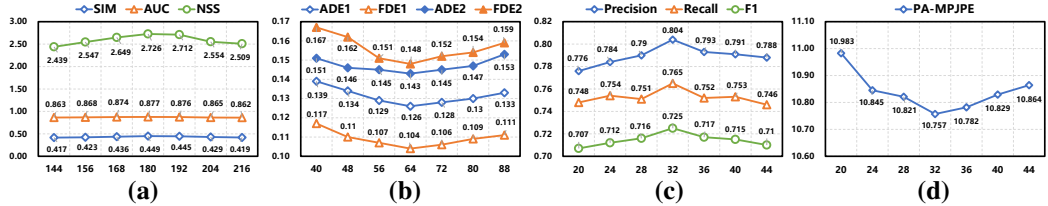

We conduct hyper-parameter experiments on the latent dimension of the C-VAE for each interaction element's decoder. The results, as depicted in Fig.~\ref{ablation}, demonstrate that the effect of each metric exhibits a trend of initially rising to a peak and then subsequently falling.
This trend occurs because the latent dimension determines the complexity of the input data in the latent space.
A smaller latent dimension results in a highly compressed input, leading to the loss of important features and adversely affecting the reconstruction results.
Conversely, a larger latent dimension increases the randomness in sampling within the latent space during inference, further increasing the uncertainty of the prediction results and reducing prediction accuracy.
Thus, these experiments demonstrate the rationality of the hyper-parameter selection in the Probabilistic Modeling Decoder, allowing the model to make accurate predictions while maintaining reasonable randomness.
\subsection{Input Study}
\begin{figure}[h]
    \centering
    \includegraphics[width=1\linewidth]{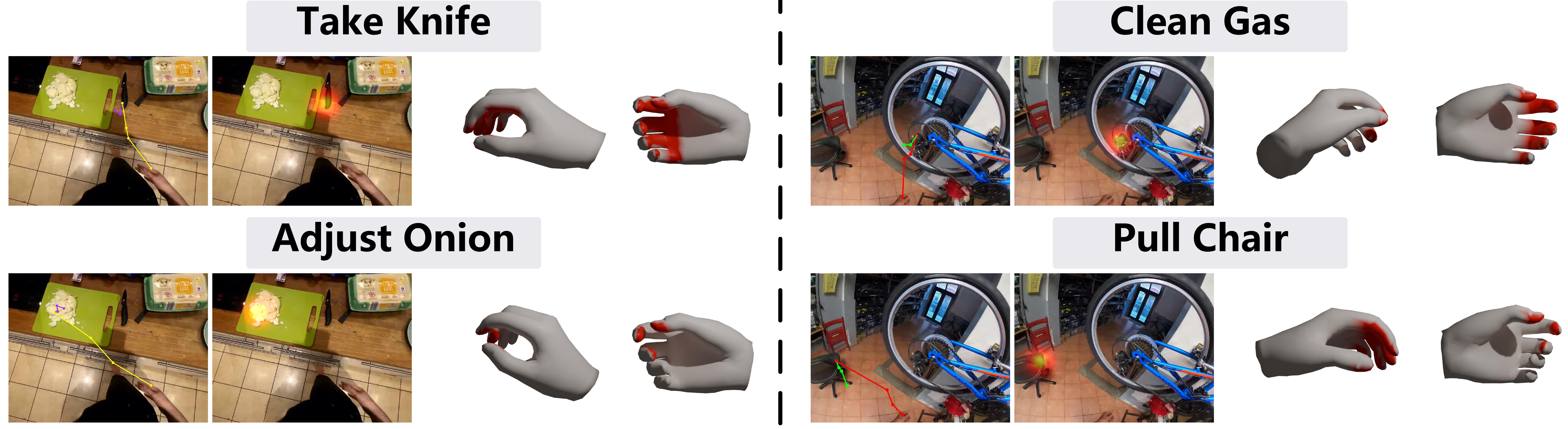}
    \caption{\textbf{Different phrases}. We fix the input image and vary the phrase to display the model's complete prediction results.}
    \label{sub1}
\end{figure}

To evaluate our model's sensitivity to variations in phrase and image inputs, we analyze experimental results by fixing one input while varying the other. 
Firstly, when the interaction scenario is held constant and the interaction prompt is altered, the model's prediction results, as shown in Fig.~\ref{sub1}, exhibit significant changes across all interaction elements. 
This demonstrates that our network is highly responsive to variations in phrases. 
It can identify different interaction objects based on various nouns and predict contact areas accordingly. 
Additionally, it can plan different hand trajectories based on different verbs, thereby achieving accurate interaction predictions through the combination of nouns and verbs.

\begin{figure}[h]
    \centering
    \includegraphics[width=1\linewidth]{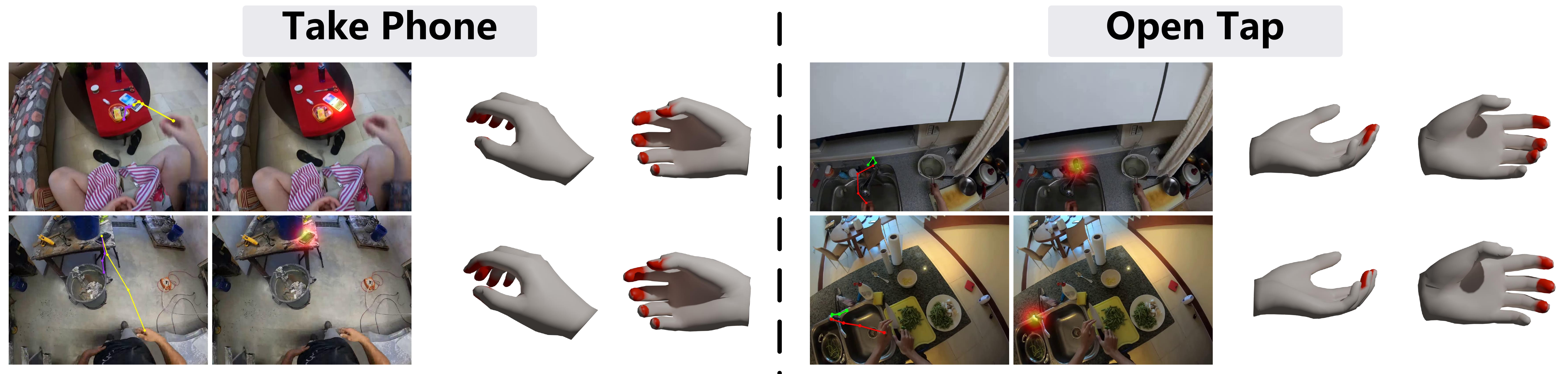}
    \caption{\textbf{Different images}. We change the images while keeping the phrase the same and display the corresponding predictions.}
    \label{sub2}
\end{figure}

Secondly, when fixing the interaction prompt and changing the interaction scenario, the model's prediction results, as illustrated in Fig.~\ref{sub2}, exhibit notable differences in interaction intentions.
This variation arises from the differing relative positions of interaction objects within the images, leading to distinct hand motion trends and interaction hotspots.
The differences in manipulation element predictions are relatively smaller.
For instance, the manipulation trajectories show certain regularities: the left sample, ``take phone'', exhibits a tendency to reverse direction, while the right sample, ``open tap'', shows an upward trend in the trajectory. 
Additionally, the predicted hand poses with contact are quite similar, demonstrating that our model can categorize hand motion patterns based on verb prompts, ensuring that the interaction manipulations predicted under the same prompt exhibit consistent regularity.

\subsection{Uncertainty Study}
\begin{figure}[h]
    \centering
    \includegraphics[width=1\linewidth]{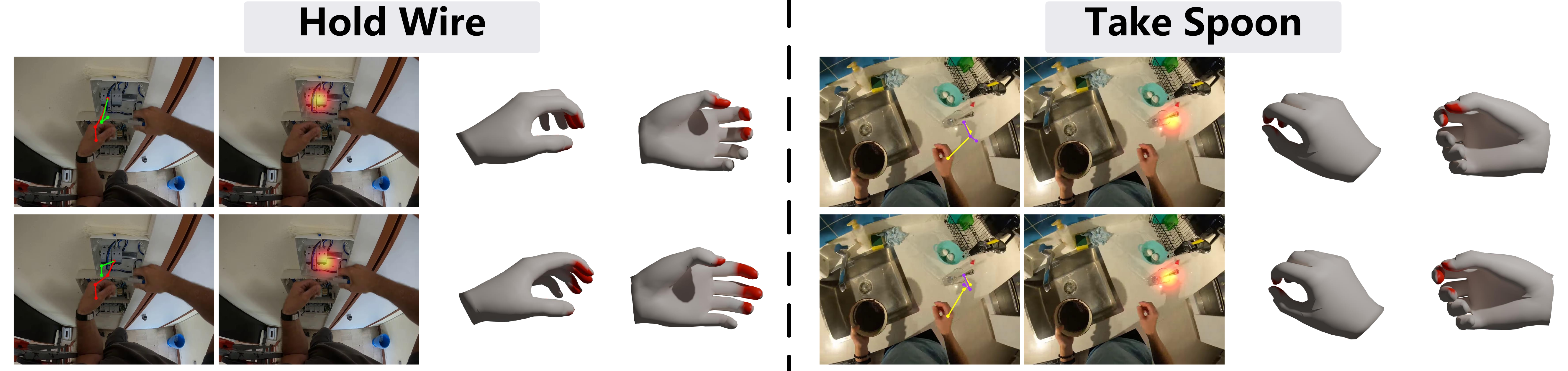}
    \caption{\textbf{Uncertainty study}. We repeat the inference with identical inputs to study the uncertainty introduced by PEAR's Probabilistic Modeling Decoder.}
    \label{fig:uncertainty}
\end{figure}

To investigate the randomness introduced by the Probabilistic Modeling Decoder in predicting outcomes, we perform repeated inferences using the same images and phrases.
The results, depicted in Fig.~\ref{fig:uncertainty}, show that the randomness in predictions is related to the specificity of the phrases and interaction scenarios.
When specificity is lower, the range of randomness is relatively larger, as illustrated on the left. For example, the term ``wire'' encompasses a certain range of area, resulting in corresponding randomness in predicting hotspots. 
However, throughout the prediction process, different interaction elements maintain consistency. When the hand pose differs from the interaction hotspots, the corresponding hand contact changes accordingly. This indicates that the established constraints ensure proper alignment between different interaction elements.
Conversely, when specificity is higher, the range of randomness is relatively smaller, as shown on the right, where the results of repeated inferences are nearly identical.
The results of the uncertainty study demonstrate that our method maintains reasonable randomness in inferences across different interaction scenarios and prompts, underscoring the robustness and rationality of the Probabilistic Modeling Decoder design.

\section{Conclusion}
In this paper, we introduce a novel model, \textbf{PEAR}, designed to anticipate comprehensive hand-object interaction. 
This anticipation involves the interaction intention prior to hand-object contact as well as the subsequent manipulation post-contact. 
To address the uncertainty in the interaction process, we decompose the phrase into verbs and nouns and utilize a cross-alignment method among verbs, nouns, and images to mitigate intention uncertainty.
Furthermore, we establish dynamic, bidirectional constraints between interaction intention and manipulation to overcome manipulation uncertainty.
To rigorously validate the proposed model's capability in hand-object interaction anticipation, we collect a new task-relevant dataset, \textbf{EGO-HOIP}, which includes both manual and automatic annotations. The PEAR model achieves state-of-the-art results on this dataset, demonstrating its superior performance.

\textbf{Future Work}. We intend to collect more precise hand-object interaction data in 3D space and conduct research on hand-object interaction understanding/anticipation within 3D environments. 
This effort will enhance the planning of interaction intention and manipulation for agents operating in 3D space.

\bibliographystyle{plain}
\bibliography{iclr2024_conference}

\end{document}